# A deep reinforcement learning platform for antibiotic discovery

Hanqun Cao[1-5, †], Marcelo D. T. Torres[2-5, †], Jingjie Zhang[1], Zijun Gao[1], Fang Wu[6], Chunbin Gu[1], Jure Leskovec[6], Yejin Choi[6], Cesar de la Fuente-Nunez[2-5,*], Guangyong Chen[7, *,] Pheng-Ann Heng[1]

[1]Department of Computer Science and Engineering, The Chinese University of Hong Kong, Hong Kong, China
[2]Machine Biology Group, Departments of Psychiatry and Microbiology, Institute for Biomedical Informatics, Institute for Translational Medicine and Therapeutics, Perelman School of Medicine, University of Pennsylvania, Philadelphia, Pennsylvania, United States of America.
[3]Departments of Bioengineering and Chemical and Biomolecular Engineering, School of Engineering and Applied Science, University of Pennsylvania, Philadelphia, Pennsylvania, United States of America.
[4]Department of Chemistry, School of Arts and Sciences, University of Pennsylvania, Philadelphia, Pennsylvania, United States of America.
[5]Penn Institute for Computational Science, University of Pennsylvania, Philadelphia, Pennsylvania, United States of America.
[6]Department of Computer Science, Stanford University, Stanford, California, United States of America
[7]Hangzhou Institute of Medicine, Chinese Academy of Sciences, Zhejiang, China.

*Correspondence:

Cesar de la Fuente-Nunez (cfuente@upenn.edu)

Guangyong Chen (chenguangyong@him.cas.cn)

# Abstract

Antimicrobial resistance (AMR) is projected to cause up to 10 million deaths annually by 2050, underscoring the urgent need for new antibiotics. Here we present ApexAmphion, a deep-learning framework for *de novo* design of antibiotics that couples a 6.4-billion-parameter protein language model with reinforcement learning. The model is first fine-tuned on curated peptide data to capture antimicrobial sequence regularities, then optimised with proximal policy optimization against a composite reward that combines predictions from a learned minimum inhibitory concentration (MIC) classifier with differentiable physicochemical objectives. *In vitro* evaluation of 100 designed peptides showed low MIC values (nanomolar range in some cases) for all candidates (100% hit rate). Moreover, 99 our of 100 compounds exhibited broad-spectrum antimicrobial activity against at least two clinically relevant bacteria. The lead molecules killed bacteria primarily by potently targeting the cytoplasmic membrane. By unifying generation, scoring and multi-objective optimization with deep reinforcement learning in a single pipeline, our approach rapidly produces diverse, potent candidates, offering a scalable route to peptide antibiotics and a platform for iterative steering toward potency and developability within hours.

**Keywords**: Deep learning, machine learning, reinforcement learning, ApexAmphion, antimicrobial peptides, peptide design, protein language models, artificial intelligence, antibiotics.

# Main

**ApexAmphion: an integrated platform for antibiotic design at scale**

AMR poses an escalating global health, with mortality and exonomic burdens expected to intensify by mid-century[1,2]. Thus, novel antibiotics are urgently needed. Artitifical intelligence approaches have been recently applied for antibiotic design and discovery[3–6]. However, reinforcement learning (RL) has not yet been applied for antibiotic design.

Peptides, including antimicrobial peptides (AMPs), constitute a promising solution to the AMR crisis due to their potent membrane-disrupting mechanisms, which are difficult for pathogens to circumvent[7]. Yet most computational discovery platforms address only parts of the pipeline—scoring[4,8–18], generation[6,19–28], or database management[12,29–32]—and are constrained by limited open-source sequences (~30,000 sequences) with inconsistent experimental annotations, especially MIC values[3,33,34].

We introduce ApexAmphion (**Fig. 1**), which integrates pretrained protein language modeling with a reinforcement learning fine-tuning stage to steer sequence generation toward potency and desirable physicochemical profiles. In stage 1, we fine-tune ProGen2-xlarge (a 6.4-billion-parameter protein language model) using supervised learning on curated AMP datasets, retaining broad protein context while capturing antibiotic-specific motifs[35,36]. In stage 2, we applied an RL procedure analogous to RL from human feedback (RLHF), where "feedback" is operationalized through expert-defined objectives: a learned MIC predictor and several physicochemical targets. Specifically, we trained ApexMIC, a binary classifier of antimicrobial activity (active defined as MIC ≤32 µmol L$^{-1}$), using ESM2-8M embeddings on a curated dataset of 7,888 positive AMP sequences (active against at least one pathogen) and 30,652 negative sequences. We then optimized the generator with proximal policy optimization (PPO), rewarding it for low ApexMIC-predicted MIC and for satisfying target ranges of charge, hydrophobicity, hydrophobic moment, length, and isoelectric point.

Post-RL, ApexAmphion samples shifted toward lower predicted MIC while remaining within ranges characteristic of effective peptide antibiotics (**Fig. 2c**). Although pathogen-specific design is not yet supported, leveraging explicit positive/negative labels across multiple datasets yielded broad-spectrum activity. *In vitro* tests on 100 ApexAmphion-generated sequences showed low MIC values for all 100 peptides, and 99 of these were active against at least two pathogens (**Fig. 3a**).

Compared to existing antimicrobial discovery approaches[4,6,12,13,21,37], ApexAmphion overcomes the limitations of current AMP data volume, achieving a high hit rate using fewer than 10,000 MIC-annotated training data points. It also greatly expands the explored antimicrobial peptide space to the order of millions of sequences via *de novo* generation. By leveraging a fine-tuned pretrained protein language model (trained on billions of natural proteins), ApexAmphion addresses the poor peptide-modeling performance of such models when used out-of-the-box, providing a strong foundation for subsequent sequence optimization. Meanwhile, fine-tuning a foundation model via RL is novel in the antibiotic discovery field, and it effectively integrates the scoring function into the generation process—transforming the traditional large-scale screening paradigm into efficient, condition-based generation. The diverse, high-quality sequences produced by the large language model successfully address the weak generalization of current small-scale models, indicating that ApexAmphion could serve as a platform for scaling up antimicrobial discovery.

ApexAmphion closes the loop between generation, *in silico* scoring, and data curation to expand high-confidence training resources. Using the platform at scale, we constructed two virtual libraries: Amphorium, containing 2.1 million non-redundant generated peptide sequences, and Amphorium-RL, a 180,000-sequence subset enriched for low predicted MIC by RL fine-tuning. Both libraries are open access and are comprehensively annotated with predicted AMP activity and MIC values (low or high) using current machine-learning tools[4,21,38,39] to facilitate downstream screening and prioritization.

**Sequence-and property-level fidelity to natural antibiotics**

Compared to other generative methods that use unsupervised training followed by AMP-specific fine-tuning, ApexAmphion (with its billions-scale evolutionary prior) more precisely captures the sequence distribution of natural AMPs (**Fig 2a, Fig 2c**). We randomly sampled 10,000 peptide sequences from ApexAmphion and from five baseline models (PepCVAE[23], HydrAMP[21], diff-AMP[24], deepAMP[25], and AMP-Designer[26]) for benchmarking. ApexAmphion demonstrated state-of-the-art performance across multiple metrics[21,23–26]. For amino acid composition (**Fig. 2a**), its samples best matched the distributions of natural AMPs, with a lower Jensen–Shannon divergence (0.142 vs 0.207 for the next best model) and higher Pearson correlation (0.851 vs 0.791) (**Fig. 2a**). ApexAmphion sequences show enriched aliphatic and hydrophobic residues (A, F, I, L, M, V) and reduced acidic/polar residues (D, E, N, Q, P) relative to natural AMPs (see **Supplementary Fig. 1** for full amino acid profiles), increasing overall cationicity and hydrophobicity consistent with known membrane-binding roles. Despite this increase in hydrophobic content, lysine (K) is enriched relative to arginine (R), favoring K-rich α-helical patterns. This indicates that the peptides maintain high charge density with greater hydration, moderating the hydrophobicity of cationic side chains and reducing nonspecific interactions with mammalian membranes (a hallmark of selective antimicrobial peptides). The model also systematically selects against glycine and proline, which are well known to disrupt secondary structure: glycine increases backbone flexibility and proline imposes rigid kinks that break α-helices. Depleting G and P biases sequences toward more continuous, stable helices, potentially facilitating membrane insertion, while a decreased frequency of cysteine reflects a preference for linear, disulfide-free scaffolds. Notably, the subset of ApexAmphion sequences that passed our in silico filters for experimental testing ("ApexAmphion-screened" peptides) showed an even stronger bias toward amphipathic composition, with further increases in V and A and further reductions in D, E, N, Q, S, T, G, and P. This yields an amphiphilic amino acid profile even more favorable for membrane interaction.

ApexAmphion samples fall in canonical AMP length ranges (8–50 residues), with a tendency toward slightly longer sequences on average than most known AMPs (**Fig. 2d and Supplementary Fig. 3a**). Mean normalized hydrophobicity values (Eisenberg scale) for the generated peptides predominantly range from 0.3 to 0.6, indicating moderate hydrophobic character suitable for membrane interaction (comparable to the diff-AMP model). Hydrophobic moment analysis (computed via helical wheel projection using the Eisenberg scale) confirms that amphipathic character is preserved (i.e., side-chain polarities are segregated), which is essential for antimicrobial activity. Net charge is typically in the +2 to +9 range, with isoelectric points above 10, enhancing electrostatic attraction to negatively charged bacterial membranes without exceeding charge levels that would risk excessive haemolysis. Importantly, the increase in cationic residues occurs primarily through lysine rather than arginine, preserving favorable hydration and selectivity. Our virtual filtering pipeline further optimized predicted hemolytic profiles, reducing cytotoxicity risk while maintaining potency: the final ApexAmphion-screened set has slightly lower average hydrophobicity and charge, trimming outliers that might be overly hydrophobic or highly charged. ESMFold predicted pLDDT scores are higher for ApexAmphion-generated peptides (especially the filtered subset) than for peptides from other models (**Supplementary Fig. 3d**), suggesting improved conformational stability and more native-like folding tendencies. Overall, ApexAmphion successfully steers candidates toward favorable selectivity–potency trade-offs. The additional filtering step narrows the variance across key parameters, indicating effective multi-objective control by the RL reward.

**Latent space similarity and diversity**

Using ESM2 embeddings of sequences, we find that ApexAmphion and its filtered subset cluster closer to the centroids of natural AMP and low-MIC peptide populations than do sequences from other generators (**Fig. 2f, g Supplementary** ). Most ApexAmphion samples lie within a small embedding distance (≤3.0) of the natural AMP manifold, and a substantial fraction are very close (distance ≤1.0), indicating high *a priori* fidelity to known AMP-like patterns. UMAP projections (**Fig. 2b** cand **Supplementary Fig. 2a-h**) show that many baseline methods collapse their outputs into limited subregions of the natural AMP space (e.g., PepCVAE, HydrAMP, diff-AMP, and AMP-Designer each populate only certain clusters of the AMP manifold), reflecting restricted diversity. Conversely, DeepAMP outputs are more dispersed but often outside major AMP regions (suggesting lower fidelity). ApexAmphion manages to capture both proximity to natural AMP distributions and broader coverage (reflected in a lower overall perplexity; (**Fig. 2e**). This likely reflects the pretrained model's ability to encode both near and distant evolutionary relationships, maintaining diversity without sacrificing realism.

We also performed a sequence similarity analysis to confirm novelty beyond the known AMP space. Using MMseqs2, we compared the Amphorium libraries to our compiled natural AMP database. The results, together with UMAP visualizations, demonstrate ApexAmphion's capability to generate truly novel sequences beyond the current AMP distribution (see diversity analysis in **Fig. 2b, h** and **Supplementary Fig. 6a, b**). For instance, a large proportion of Amphorium peptides share <70% identity with any known AMP, and the generative model contributes many new sequence clusters that are not present in natural databases. This indicates that generative modeling and genome/metagenome mining are accessing complementary regions of peptide sequence space, expanding the discoverable antimicrobial landscape beyond what either approach alone could cover.

**Amphorium: a 2.1-million-sequence virtual library expanding peptide antibiotic space**

The exponential growth of drug-resistant infections necessitates innovative discovery approaches that transcend traditional compound libraries. We used ApexAmphion's two-stage generative pipeline to construct unprecedented virtual libraries of antimicrobial peptide candidates *de novo*. The ApexAmphion-SFT model (after supervised fine-tuning) was used to sample a broad diversity of sequences, and the ApexAmphion-RL model (after reinforcement learning optimization) was used to generate sequences biased toward low predicted MIC. All generated sequences underwent rigorous quality control, including length filtering (retaining sequences 8–50 amino acids long) and removal of exact duplicates, followed by comprehensive annotation with predicted activities and properties. In particular, we applied our ApexMIC classifier as well as external tools (e.g., APEX[4], HydrAMP[21], AMPScanner2[39], and physicochemical property analysis[40]) to each sequence, and we computed key physicochemical descriptors. This systematic approach yielded two complementary resources: Amphorium (2.1 million non-redundant peptide sequences from the supervised model) and Amphorium-RL (180,000 sequences from the RL-optimized model), both annotated with AMP classification and MIC prediction (active vs inactive at 32 µmol L$^{-1}$ threshold).

**Amphorium composition and physicochemical properties**

Amphorium closely matches natural AMP composition. The per-residue frequencies of most amino acids differ by less than 0.01 from those of natural AMPs (Jensen–Shannon distance = 0.045; Pearson r = 0.97) (**Fig. 2a** and **Supplementary Fig. 3a**). Amphorium-RL, in contrast, shows intentional biases: for example, lysine frequency increases by +0.088, and both glycine and leucine increase by >0.03. These enrichments suggest the RL model favors simpler, flexible helical sequences with an enhanced propensity for membrane interaction. Correspondingly, Amphorium-RL shows marked decreases in glutamic acid (–0.030) and in several other polar or acidic residues (D, Q, S, T each reduced by >0.02), which would strengthen electrostatic and hydrophobic interactions with negatively charged bacterial membranes. The overall physicochemical property distributions of Amphorium-RL mirror those observed for the ApexAmphion filtered-screened subset described earlier: namely, higher average charge/pI and hydrophobic moment, moderate hydrophobicity, and a trend toward sequences in the mid-length range of known AMPs.

**Amphorium-RL latent space structure**

In the ESM2 embedding space, Amphorium-RL occupies a biased subregion within the broader Amphorium distribution, consistent with the reward-driven selection of candidates with enhanced predicted potency (i.e., lower MIC). At the same time, latent space analysis shows that Amphorium (the large diverse set) and AMPSphere[12] (a database of mined natural AMPs) collectively span the known AMP manifold while also occupying distinct regions beyond it. In other words, generative modeling and genome/mining approaches retrieve largely non-overlapping novel sequences. Together, they cover the landscape of known AMPs and extend into new territories that either approach alone would miss. This highlights their complementarity: generative methods like ApexAmphion can explore sequence space unconstrained by nature, while mining methods can find peptides arising from evolution, and each uncovers candidates the other might overlook.

**Virtual library value for candidate triage**

We evaluated the Amphorium libraries versus AMPSphere using pathogen-specific MIC prediction models (APEX 1.1[4,17]; results summarized schematically in **Supplementary Figs. 7-9**). Overall, Amphorium and AMPSphere show similar predicted MIC distributions across 11 important pathogens, with the majority of candidates predicted to have MICs above 200 µmol L$^{-1}$. Amphorium-RL, however, is substantially enriched in potent candidates: its MIC distributions are shifted downward, with the lower quartile below 128 µmol L$^{-1}$ for multiple pathogens (e.g., *Acinetobacter baumannii* ATCC 19606; *Escherichia coli* AIC221 and polymyxin-resistant AIC222; and vancomycin-resistant *Enterococcus faecium* ATCC 700221). Using a stringent activity threshold of 32 µmol L$^{-1}$, the "pass rates" (fraction of peptides predicted to be active) for Amphorium and AMPSphere are modest overall, but they are consistently higher for pathogens with more abundant training data (e.g., *E. faecium* VRE, *A. baumannii*, several *E. coli* strains, *Pseudomonas aeruginosa* PAO1/PA14). This suggests a larger discoverable space for well-represented pathogens. Notably, Amphorium produces far more absolute candidates than AMPSphere for each pathogen—e.g., ~100,000 predicted actives for VRE *E. faecium*, versus ~10,000 from AMPSphere—amounting to 404,201 pathogen-specific "hits" in Amphorium (cumulative across pathogens) versus 34,275 in AMPSphere. Amphorium-RL further boosts both the hit rates and absolute counts (achieving ~10% predicted actives for some major pathogens), yielding a total of 91,358 predicted hits in this enriched subset.

This trend holds across independent scoring methods: more than 50% of Amphorium-RL sequences and 22% of Amphorium sequences pass the HydrAMP low-MIC filter (versus ~14% for AMPSphere). Using AMPScannerv2, the Amphorium libraries likewise show substantially higher AMP probability scores and pass rates than AMPSphere. These comparisons underscore that ApexAmphion's generative libraries can provide a rich pool of high-confidence candidates to improve screening efficiency.

**Amphorium enhances the efficiency of antimicrobial discovery**

Mining microbial genomes and metagenomes often entails triaging billions of putative small open reading frames (smORFs) or peptides, many of which lie outside natural amino acid distributions or are unsuitable for synthesis and development. Scoring functions trained on natural proteins can over-score such out-of-distribution sequences, yielding unstable or non-synthesizable false positives. Consequently, reported hit rates in purely genome-mining pipelines can be <0.1%. By contrast, Amphorium and Amphorium-RL offer focused, richly annotated sets of candidates that improve downstream screening efficiency. They can also serve as "virtual data" to expand model training and to guide targeted experimental validation by highlighting peptides that satisfy multiple design criteria.

**Antimicrobial activity of amphionins against bacterial pathogens**

To validate ApexAmphion's predictions, we synthesized and tested 100 amphionins *in vitro* against a panel of pathogenic bacteria. This panel included six Gram-negative species (*Acinetobacter baumannii*, *Escherichia coli*, *Klebsiella pneumoniae*, *Pseudomonas aeruginosa*, *Salmonella enterica*, *Enterobacter cloacae*) and four Gram-positive species (*Staphylococcus aureus*, *Bacillus subtilis*, *Enterococcus faecalis*, *Enterococcus faecium*), encompassing both drug-susceptible strains and multidrug-resistant clinical isolates. All 100 amphionins inhibited bacterial growth at concentrations ≤64 µmol L$^{-1}$, achieving a 100% hit rate (**Fig. 3a**). Moreover, 99 of the 100 peptides were active against two or more different pathogens, indicating broad-spectrum efficacy.

Potency was pronounced across the Gram-negatives. For example, both drug-susceptible and multidrug-resistant *A. baumannii* (the resistant strain is non-susceptible to ceftazidime, gentamicin, ticarcillin, piperacillin, aztreonam, cefepime, ciprofloxacin, imipenem, and meropenem) showed median MICs of 2-4 µmol L$^{-1}$. Several *E. coli* strains—including a polymyxin/colistin-resistant isolate—were inhibited at 2-8 µmol L$^{-1}$. *P. aeruginosa* PAO1 had a median MIC of ~8 µmol L$^{-1}$. Gram-positive pathogens were similarly susceptible: for instance, methicillin-resistant *S. aureus* (MRSA) and vancomycin-resistant *E. faecium* both had MICs in the 8-16 µmol L$^{-1}$ range. These values rival or surpass those of conventional antibiotics tested in the same assays, underscoring the potency of the amphionin peptides.

Analyzing sequence features of the most active amphionins (those with lowest MICs) reveals clear trends consistent with our design objectives. Potent sequences carry a high net positive charge (+4 to +7) primarily due to lysine-rich content, providing strong electrostatic attraction to negatively charged bacterial membranes while avoiding the excessive hydrophobicity and potential toxicity of arginine-rich peptides. The amphionins are also enriched in hydrophobic residues (especially L, F, V, A, M), which facilitate membrane partitioning and insertion. Conversely, they are strongly depleted in acidic residues (D, E) and polar uncharged residues (N, Q, S, T), minimizing hydrophilic interactions and biasing the peptides toward membrane affinity. Furthermore, the most active peptides have low proportions of glycine and proline, which is consistent with maintaining secondary structure for effective membrane disruption. Cysteine is almost completely absent in these sequences, indicating a preference for linear, non-disulfide-bonded peptides that are easier to synthesize and do not depend on oxidative folding.

Together, these design features help explain the remarkable activities observed. The amphionins combine: (i) high cationicity through lysine-rich motifs; (ii) well-balanced amphiphilicity (sufficient hydrophobic content to disrupt membranes, but with polar residues curtailed); (iii) avoidance of secondary structure-breaking residues; and (iv) linear, flexible scaffolds. The result is a library of synthetic peptides with broad-spectrum, low-MIC activity against both drug-susceptible and drug-resistant bacteria.

**Membrane-disruptive mechanism of action of amphionins.** To elucidate how amphionins kill bacteria, we examined their effects on bacterial membranes using fluorescence assays in *A. baumannii* ATCC 19606. We monitored outer membrane permeability with the NPN uptake assay [1-(N-phenyl-naphthylamine) becomes fluorescent in a hydrophobic environment], and we measured cytoplasmic membrane depolarization with the 3,3′-dipropylthiadicarbocyanine iodide ($DiSC_3$-5) assay, which detects loss of transmembrane polarization. In these assays, the amphionins showed clear signatures of membrane disruption at their active concentrations, with a predominant effect on the cytoplasmic membrane (**Fig. 3b**). We used Triton X-100 as a positive control (maximal membrane lysis) and included polymyxin B (a membrane-acting peptide antibiotic) and levofloxacin (a DNA-targeting antibiotic) for comparison.

When comparing peptides by their maximum induced fluorescence, only a subset of amphionins caused strong outer membrane permeabilization (e.g., amphionin-23, -43, -46, -51, -60 showed high NPN uptake peaks). In contrast, many more amphionins induced potent cytoplasmic membrane depolarization (e.g., amphionin-38, -42, -46, -49, -51, -63 showed high $DiSC_3$-5 peaks). This indicates that collapsing the inner membrane potential is the dominant mechanism for most amphionins, whereas outer membrane disruption is less common and appears to be sequence-specific (**Fig. 3b–d**).

Analyzing the kinetics and duration of the membrane effects provided further insight. We considered both the peak magnitude of each fluorescence signal and the area under the curve (AUC) over time. Some amphionins (e.g., amphionin-38, -42,- 49) achieved high depolarization peaks and large AUCs in the $DiSC_3$-5 assay, meaning they rapidly and sustainably collapsed the membrane potential. Others (e.g., amphionin-46, and -63) had high peaks but smaller AUCs, suggesting a potent but more transient depolarization. A few peptides showed moderate depolarization peaks yet accumulated large AUCs, indicating a slower but persistent disruptive effect. In the NPN assay, a similar dichotomy was observed: a small number of amphionins caused sharp but brief outer membrane permeabilization, whereas others showed gradual, sustained permeabilization with lower peak intensity. Notably, there was little correlation between a peptide's NPN response and its DiSC$_3$-5 response (**Fig. 3c, d**), reinforcing that outer membrane perturbation and inner membrane depolarization are largely independent properties among this peptide set.

Sequence-level differences help explain these mechanistic classes. The strongest depolarizers (those with high DiSC$_3$-5 peak and AUC) tended to be slightly longer (~16-20 residues), with net charges of +6 or greater, and they maintained very lysine-rich sequences. These peptides often included aromatic residues like phenylalanine or tyrosine, which can facilitate deeper insertion into the lipid bilayer and stabilize peptide–membrane interactions. In contrast, amphionins that caused more pronounced outer membrane permeabilization (high NPN responders) were typically shorter and less charged, often featuring motifs rich in leucine and serine that may favor a more superficial binding to the outer membrane lipids without deep penetration. Across both groups, effective amphionins consistently minimized secondary structure-disrupting residues (G, P) and lacked cysteines, thus remaining largely linear — traits conducive to membrane interaction.

In summary, amphionins appear to kill bacteria primarily by targeting the cytoplasmic membrane, causing rapid depolarization and loss of membrane potential. Outer membrane permeabilization occurs with certain peptides but is not a prerequisite for activity against *A. baumannii* (which is consistent with polymyxin-like behavior, where some peptides may transit the outer membrane via self-promoted uptake). By analyzing both peak effects and temporal dynamics, we identified two complementary mechanistic profiles in the amphionin library: "rapid inserters" that cause immediate, intense disruptions, and "steady disruptors" that cause sustained membrane stress. Both profiles emerged from the RL-guided design, illustrating how ApexAmphion converged on different membrane-targeting strategies through sequence optimization.

**Discussion**

We have presented ApexAmphion, a scalable platform that leverages pretrained protein language models and reinforcement learning to transform antimicrobial discovery. Our approach harnessed limited and heterogeneous public AMP data to generate and evaluate millions of candidate sequences, representing a significant advance in computational antibiotic discovery. Through comprehensive experimental validation of 100 designed peptides — including MIC testing and mechanistic assays — we demonstrated that large-scale modeling can overcome the distributional biases and limited diversity that constrained previous peptide generation efforts. ApexAmphion thereby enabled AMP discovery at an unprecedented scale.

The superior fidelity and diversity achieved by our model stem fundamentally from leveraging the protein "universe" distribution as an informative prior. Current computational approaches typically train relatively small generative models only on the thousands of known human-discovered AMPs, attempting to capture the peptide distribution from this narrow sample. This scarcity of training data leads to biased models and limited generative diversity. In contrast, large protein language models trained on billions of natural sequences possess extensive general knowledge, but they are not specifically tuned to short antimicrobial peptides and thus perform suboptimally on this task out-of-the-box. We addressed this by fine-tuning ProGen2-XL (the largest available open-source protein generator) on AMP data using a lightweight Low-Rank Adaptation (LoRA) strategy. This strategy avoids catastrophic forgetting of general protein knowledge while successfully specializing the model to produce high-fidelity AMP-like sequences with maintained diversity.

ApexAmphion outperforms prior approaches by substantial margins on computational metrics. Its generated sequences have amino acid compositions, physicochemical property distributions, and embedding distances that are closest to those of natural AMPs. At the same time, diversity analyses (e.g., UMAP projections) confirm that ApexAmphion samples are not mere copies of known AMPs but cover a broad and novel sequence space, with precise control over both fidelity and diversity. Building on these capabilities, we constructed the Amphorium database — a comprehensive virtual repository of over 2 million machine-generated AMP candidates, clustered into more than 1 million sequence families. By annotating Amphorium with multiple state-of-the-art predictors, we showed that it provides a far richer pool of leads than AMPSphere's ~800,000 entries obtained from microbiomes. Finally, experimental validation yielded a 100% success rate (all 100 tested peptides showed activity), conclusively demonstrating the advantage of incorporating broader priors and multi-objective optimization into AMP design.

Despite its success, the ApexAmphion framework has limitations. Currently, Amphion's generative criteria are constrained to a binary classification of "high" vs "low" antimicrobial activity, rather than producing peptides tailored to specific pathogens or conditions. This is due to the limited availability of consistent multi-pathogen MIC values datasets for training. Additionally, heterogeneity in how MIC values data were measured across different sources likely impacts the accuracy of ApexMIC and thus limits the quality of the reward signal; this in turn may cause the model to miss some high-potential candidates or to favor sequences that align with experimental biases.

Future developments will seek to expand ApexAmphion's scope and controllability. With more data becoming available, we anticipate implementing pathogen-specific or species-targeted generation, as well as incorporating additional design constraints (for example, tuning peptides for certain secondary structures, or minimizing immunogenic motifs). We also plan to refine our reward functions and integrate more advanced property predictors (for stability, protease resistance, etc.), which could further improve the developability of generated peptides. The introduction of resources like Amphorium and models like ApexMIC will hopefully catalyze new methodological advances, better screening approaches, and accelerated discovery pipelines in the AMP field.

In summary, the ApexAmphion platform represents a significant step toward addressing the challenge of antimicrobial resistance. By uniting large-scale protein knowledge with task-specific refinement and reinforcement learning, it achieves both breadth and precision in peptide discovery. This work lays the groundwork for next-generation antibiotic development and illustrates the promise of AI-driven bioengineering for tackling urgent global health threats.

## Methods

### Supervised fine-tuning

We employed ProGen2-xlarge (6.4 billion parameters), a GPT-like model, as the base sequence generator due to its flexible autoregressive design and rich pretraining on diverse proteins. To efficiently specialize this model for AMPs, we used Low-Rank Adaptation (LoRA), which adds a small number of trainable parameters to each transformer layer. This approach allows fine-tuning on the AMP task without overfitting or forgetting the general protein language. We fine-tuned the model on our curated AMP dataset (see Data Preparation) for several epochs, optimizing the standard next-amino-acid prediction loss. Perplexity on a validation set was used to guide training and prevent over-training, ensuring the generated sequences remained similar to the training distribution in functionally relevant ways.

The supervised fine-tuning loss function is defined as:

$$L_{SFT} = -\sum_{i=1}^{N}\sum_{t=1}^{T} logP(x_{i,t}|x_{i,<t}; \theta_{LoRA})$$

where N is the number of sequences in the training batch; T is the length of each sequence; $x_{i,t}$ represents the t-th token of the i-th sequence; $x_{i,<t}$ denotes all tokens before t in the i-th sequence; and $\theta_{LoRA}$ represents the LoRA parameters.

### Reward function design

To guide the generator toward potent and well-behaved peptides, we designed a composite reward that combines a learned MIC predictor with multiple property-based objectives.

### MIC predictor (ApexMIC)

We developed ApexMIC to evaluate the likelihood that a given peptide has strong antimicrobial activity (operationally defined as MIC ⩽32 μmol L$^{-1}$). Peptide sequences were encoded using ESM2 (8 million parameter version) to obtain feature vectors, which were input to a multi-layer perceptron that outputs an activity score. The model was trained on a labeled dataset of AMPs with known activity (see Data Preparation) using a focal loss to handle class imbalance.

To address the class imbalance commonly observed in antimicrobial activity datasets, we utilize Focal Loss as our training objective:

$$L_{focal} = -\frac{1}{N}\sum_{i=1}^{N}\alpha_i(1-p_i)^\gamma \log(p_i)$$

where $p_i$ is the predicted probability for the true class of the i-th sample; $\alpha_i$ is the weighting factor for class imbalance; γ is the focusing parameter that down-weights easy examples; N is the number of training samples. This loss emphasizes learning from the harder, informative examples (e.g., borderline activity peptides). The trained ApexMIC model outputs a score *s* in [0,1] for each peptide (higher = more likely to have MIC ≤32 μmol L$^{-1}$). For RL, we converted this into a reward component $R_{MIC}$ that encourages high *s*. Specifically, we set a target of *s* = 0.4 (the classifier's decision boundary) and defined $R_{MIC}$ such that a peptide gets a positive reward if *s > 0.4* and a negative "penalty" if *s < 0.4*.

$$R_{MIC} = \begin{cases} (s - \gamma) * \beta, & s < 0.5 \\ 1.0, & s \geq 0.5 \end{cases}$$

We applied a scaling factor β = 4 and an offset γ = 0.35 to avoid gradient instability (these values were tuned empirically). In effect, peptides confidently predicted to be active receive a strongly positive reward, those predicted inactive get negative reward, and those near the margin get a smaller signal to prevent oscillations.

**Physicochemical reward**

We included five key peptide descriptors in the reward: hydrophobicity, hydrophobic moment, net charge, isoelectric point (pI), and sequence length. These properties were computed as follows:

**Hydrophobicity** was calculated using the Eisenberg hydrophobicity scale, which assigns hydrophobicity values to individual amino acids based on their transfer free energy from water to organic solvents. The global hydrophobicity of each peptide was computed as the arithmetic mean of individual amino acid hydrophobicity values.

**Hydrophobic moment** was determined using the Eisenberg scale in conjunction with the helical wheel projection method. This parameter quantifies the amphiphilicity of peptides by measuring the magnitude of the hydrophobic moment vector when amino acids are arranged in an idealized α-helical conformation.

**Net charge** was calculated at physiological pH (7.0) by summing the charges of ionizable amino acids. Positively charged residues (Lys, Arg, His) contributed +1 each, while negatively charged residues (Asp, Glu) contributed -1 each.

**Isoelectric point (pI)** was computed as the pH at which the peptide carries no net charge, determined by iteratively solving the Henderson-Hasselbalch equation for all ionizable groups in the sequence.

The property-based reward component is formulated as:

$$R_{property} = d_1 * clamp(p_h, -0.5, 0.8) + d_2 * clamp(p_{hm}, 0.0, 0.6) + d_3 \\ * clamp(p_q, -5.0, 9.0) + d_4 * clamp(p_{is}, 8.0, 11.0) + d_5$$

where $p_h$, $p_{hm}$, $p_q$, and $p_{is}$ represent hydrophobicity, hydrophobic moment, charge, and isoelectric point respectively. The clamp function constrains these properties within specified ranges to prevent the generation of AMPs with undesirable characteristics.

**Integrated reward**

The total reward R for RL was a weighted sum of the MIC-based reward and the property reward:

$$R_{total} = \lambda * R_{property} + (1 - \lambda) * R_{MIC}$$

We chose λ (=0.5) to give roughly equal emphasis to maintaining good properties and achieving high predicted potency.

This integrated approach enables simultaneous optimization of both structural feasibility (through physicochemical constraints) and functional efficacy (through MIC prediction), providing a comprehensive reward signal for generating high-quality AMPs.

**Reinforcement learning fine-tuning**

In stage 2, we fine-tuned the ApexAmphion generator using RL (specifically, PPO) to maximize the integrated reward described above. The policy model was initialized from the supervised fine-tuned model (we refer to this as Amphion-SFT). Only the LoRA adapter parameters remained trainable, keeping the number of updated parameters small. We generated peptide sequences (modeled as trajectories of amino acid "actions") using the current policy and evaluated each with the reward function $R$. The advantage of each action was estimated via a value network (also a LoRA-equipped ProGen2 model head). We optimized the policy with the PPO objective:

**Loss Function**

We employ the Proximal Policy Optimization (PPO) algorithm to fine-tune the protein language model using the integrated reward function. The PPO loss function is expressed as:

$$L_{PPO} = L_{policy} + c_1 L_{value} - c_2 H(\pi_\theta)$$

where: $L_{policy}$ is the policy loss; $L_{value}$ is the value function loss; $H(\pi_\theta)$ is the policy entropy, and $\pi_\theta$ denotes the policy—Amphion-SFT where only the same LoRA parameters $\theta_{LoRA}$ are trainable; $c_1$ and $c_2$ are weight coefficients.

The components of this loss function are:

$$L_{policy} = -\frac{1}{NM} \sum_{i=1}^{N} \sum_{j=1}^{M} \min\left(r_{ij}(\theta)\, \hat{A}_{ij},\, clip(r_{ij}(\theta), 1 - \epsilon, 1 + \epsilon)\, \hat{A}_{ij}\right)$$

$$L_{value} = \frac{1}{NM} \sum_{i=1}^{N} \sum_{j=1}^{N} \left(V_\theta(s_{ij}) - \bar{R}_{ij}\right)^2$$

$$H(\pi_\theta) = -\frac{1}{NM} \sum_{i=1}^{N} \sum_{j=1}^{N} \sum_{k} \pi_\theta(a_k|s_{ij}) \log \pi_\theta(a_k|s_{ij})$$

where N denotes the number of peptides generated in each batch and M denotes the number of rollout steps. $\theta$ denotes the parameters of policy model; $r_{ij}(\theta)$ denoted the probability ratio for the action taken at timestep j by actor i, which is calculated below; $\hat{A}_{ij}$ denotes the estimated Advantage Function at timestep j for actor i. It quantifies how much better a specific action was compared to the average action at that state. $\epsilon$ denotes the threshold for probability ratio clip; $V_\theta$ represents the value model, $s_{ij}$ represents the state (generated peptide) at timestep j by actor i, and $a_k$ denotes the k-th action on j-th residue of the generate peptide $s_{ij}$. $\bar{R}_{ij}$ denotes the predicted score of ApexMIC of $s_{ij}$.

$$r_{ij}(\theta) = \frac{\pi_{theta}(a_{ij}|s_{ij})}{\pi_{\theta_{old}}(a_{ij}|s_{ij})}$$

To enhance training stability, we apply reward processing including scaling, normalization, and whitening below:

$$\tilde{R}_{ij} = \frac{R_{ij}}{\max(R_{ij})} \quad \bar{R}_{ij} = \frac{\tilde{R}_{ij} - \mu_{\tilde{R}}}{\sigma_{\tilde{R}}}$$

Where $R_{ij}$ represents the integrated reward $R_{total}$ for sequence $s_{ij}$; $\mu_R$ and $\sigma_R$ denotes the mean and standard deviation of all $R_{ij}$. $\tilde{R}_{ij}, \bar{R}_{ij}$ denote the scaled and normalized scaled reward respectively. The detailed training settings are provided in the Supplementary Materials.

**Virtual screening and candidate selection**

Following peptide generation by ApexAmphion, , we implemented a multi-stage virtual screening pipeline to identify the most promising peptide candidates for synthesis.

**ApexMIC screening:** The first filter was our ApexMIC predictor. We evaluated every generated sequence with ApexMIC and selected those above a probability cutoff (we used 0.4, slightly below the 0.5 decision boundary, to be inclusive while still enriching for likely actives). Sequences predicted to be inactive (score below 0.4) were discarded. This step reduced the library to peptides with a high chance of low MIC activity. We also eliminated any sequences predicted to be extremely insoluble or difficult to synthesize (e.g., very hydrophobic sequences, or containing motifs prone to aggregation or cyclization).

**Structural and physicochemical filtering:** Next, we applied several heuristic filters to ensure selected peptides were suitable for synthesis and testing. We limited peptide length to ⩽50 amino acids to keep chemical synthesis feasible. We evaluated folding stability and secondary structure using predictive tools (AlphaFold2 and ESMFold) — candidates predicted to form complex tertiary structures (e.g., requiring disulfide bonds or likely to misfold) were de-prioritized in favor of those predicted to be primarily linear and flexible (since our design mechanism of action is membrane disruption by relatively unstructured amphipathic helices). We also assessed each peptide's similarity to known proteins to avoid anything highly homologous to human proteins (which could pose toxicity or immunogenicity concerns). Specifically, we used MMseqs2 to search UniRef50[41,42] for each peptide, and we filtered out any peptide with a significant match covering >70% of its length at high identity (**Table 1** shows the few marginal hits that were found). The vast majority of amphionins had no close matches in UniRef, confirming their novelty.

**Candidate prioritization:** Finally, we integrated the above analyses to prioritize peptides for experimental validation. We favored peptides that (i) had high ApexMIC scores (and also scored well on external models like HydrAMP and AMPScanner), (ii) satisfied ideal property criteria (moderate hydrophobicity, strong amphipathicity, charge in a good range, low predicted hemolysis), (iii) showed stable predicted helicity and lack of problematic motifs, and (iv) were sufficiently novel (i.e., not essentially identical to a known AMP). We also gave consideration to diversity — selecting a set of 100 peptides that covered a range of sequence patterns rather than many near-duplicates. Peptides that met all criteria were finalized for synthesis and testing.

**Data Prepartion**

Two distinct datasets were compiled from open-source databases to support the model training framework. The first dataset comprised naturally occurring antimicrobial peptides (AMPs) and was utilized for supervised fine-tuning to align ProGen2's learned distribution with the characteristic features of antimicrobial peptides. The second dataset incorporated minimum inhibitory concentration (MIC) values, strategically augmented with high-MIC samples serving as negative training examples, and was specifically designed for ApexMIC model optimization.

**Dataset for supervised fine-tuning.** To comprehensively explore the generative capacity of protein language models, we compiled a diverse dataset of AMPs from multiple open-source databases, including DRAMP, DADP, LAMP2, dbAMPv2.0, and DBAASP[29,30,43–45]. We focused on AMPs with reported antimicrobial, antibacterial, and antifungal activities, as well as those with known minimum inhibitory concentration (MIC) values. After filtering for length (<50 amino acids) and removing redundancies, our final dataset comprised 27,148 unique sequences. We derived training and testing sets based on MMSeqs2 clustering analysis[42]. To facilitate efficient fine-tuning, we categorized the AMPs into subfamilies based on their functional annotations. This curated dataset underpins our subsequent fine-tuning and machine learning-based filtering steps, enabling a comprehensive exploration of the natural AMP distribution.

**Dataset for feedback (RL) fine-tuning.** To construct a high-quality dataset of low Minimum Inhibitory Concentration (MIC) antimicrobial peptides (AMPs), we aggregate sequences from five comprehensive public databases: DRAMP, CAMP-R4, APD3, DADP, and LAMP2[30,31,38,43,44]. Positive samples are rigorously selected based on their antibacterial, antifungal, anti-gram-positive, or anti-gram-negative activity, with MIC values ≤32 μM/mL and peptide lengths ranging from 12 to 50 amino acids. Only experimentally validated sequences are included, and redundancies are removed. The negative dataset is assembled from two sources: the Veltri negative dataset[39], containing experimentally verified non-low MIC AMPs, and a computationally augmented set derived from UniRef data and collected AMP data with unverified MIC result classified using the HydrAMP MIC classifier with a stringent threshold of 0.01[21].

Negative samples underwent length filtering (8-50 amino acids) and are deduplicated using CD-Hit with a 40% similarity threshold. To mitigate potential biases, we balance the positive and negative sample distributions and ensured similar length distributions between the two classes. A valid amino acid filter is applied to enhance dataset quality. The final dataset, comprising 38,623 sequences, is partitioned into training (30,914), validation (3,853), and test (3,856) sets. This meticulous preparation process yield a robust, balanced, and representative low MIC AMP dataset, suitable for advanced machine learning model development and evaluation in antimicrobial peptide research.

**Peptide Synthesis**

The 100 selected amphionin peptides were synthesized by solid-phase peptide synthesis (AAPPTec) using standard Fmoc (9-fluorenylmethoxycarbonyl) chemistry. Cleavage and deprotection were performed with appropriate cocktails, and crude peptides were precipitated and lyophilized. Each peptide was purified (if necessary) and verified by analytical reverse-phase HPLC and MALDI-TOF mass spectrometry. Peptide purity was >95% for all sequences. Lyophilized peptides were stored desiccated at –20 °C and reconstituted in sterile water or buffer immediately before use in assays.

**Culturing conditions and bacterial strains**

The pathogenic strains utilized included *Acinetobacter baumannii* ATCC 19606, *Acinetobacter baumannii* ATCC BAA-1605 (resistant to ceftazidime, gentamicin, ticarcillin, piperacillin, aztreonam, cefepime, ciprofloxacin, imipenem, and meropenem), *Escherichia coli* ATCC 11775, *Escherichia coli* AIC221 [MG1655 phnE_2::FRT, polymyxin-sensitive control], *E. coli* AIC222 [MG1655 pmrA53 phnE_2::FRT, polymyxin-resistant], *Escherichia coli* ATCC BAA-3170 (resistant to colistin and polymyxin B), *Enterobacter cloacae* ATCC 13047, *Klebsiella pneumoniae* ATCC 13883, *Klebsiella pneumoniae* ATCC BAA-2342 (resistant to ertapenem and imipenem), *Pseudomonas aeruginosa* PAO1, *Pseudomonas aeruginosa* PA14, *Pseudomonas* *aeruginosa* ATCC BAA-3197 (resistant to fluoroquinolones, beta-lactams, and carbapenems), *Salmonella enterica* ATCC 9150, *Salmonella enterica* subsp. *enterica* Typhimurium ATCC 700720, *Bacillus subtilis* ATCC 23857, *Staphylococcus aureus* ATCC 12600, *Staphylococcus aureus* ATCC BAA-1556 (resistant to methicillin), *Enterococcus faecalis* ATCC 700802 (resistant to vancomycin), and *Enterococcus faecium* ATCC 700221 (resistant to vancomycin). *Pseudomonas* isolates were cultured on selective Pseudomonas Isolation Agar. All other bacteria were propagated using LB (Luria-Bertani) agar and broth. Each culture was initiated from a single colony, incubated overnight at 37 °C, and subsequently diluted 1:100 into fresh media to grow to mid-log phase.

**Minimal inhibitory concentration (MIC) determination**

MICs were determined by the broth microdilution method in 96-well plates, following CLSI guidelines with slight modifications for peptides. Each amphionin peptide was tested in Mueller-Hinton Broth (for consistency with standard antibiotic testing) or LB broth as specified. Peptides were two-fold serially diluted in sterile water across the plate (final concentration range 0.78 µmol $L^{-1}$ to 64 µmol $L^{-1}$ after inoculation). Mid-log phase bacteria were diluted to ~$4 \times 10^6$ CFU $mL^{-1}$ in broth, and 50 µL of this inoculum was added to 50 µL of peptide solution in each well (resulting in ~$2 \times 10^6$ CFU $mL^{-1}$ and the desired peptide concentrations). Growth controls (no peptide) and sterile blanks were included on each plate. Plates were incubated at 37 °C for 18-20 h and then read visually and by optical density at 600 nm. The MIC was defined as the lowest peptide concentration at which no visible growth was observed ($OD_{600}$ ~ background). All MIC assays were performed in triplicate on separate days. For quality control, reference antibiotics (e.g., polymyxin B for Gram-negatives, vancomycin for Gram-positives) were tested in parallel against representative strains to ensure expected MIC ranges.

**Outer membrane permeabilization assays**

N-phenyl-1-napthylamine (NPN) uptake assay was used to evaluate the ability of the peptides to permeabilize the bacterial outer membrane. Inocula of *A. baumannii* ATCC 19606 were grown to an OD at 600 nm of 0.4 $mL^{-1}$, centrifuged (9,391 ×g at 4 °C for 10 min), washed and resuspended in 5 mmol $L^{-1}$ HEPES buffer (pH 7.4) containing 5 mmol $L^{-1}$ glucose. The bacterial solution was added to a white 96-well plate (100 µL per well) together with 4 µL of NPN at 0.5 mmol $L^{-1}$. Consequently, peptides diluted in water were added to each well, and the fluorescence was measured at $\lambda_{ex}$ = 350 nm and $\lambda_{em}$ = 420 nm over time for 45 min. The relative fluorescence was calculated using the untreated control (buffer + bacteria + fluorescent dye) as baseline and the following equation was applied to reflect % of difference between the baselines and the sample:

$$Percentage\ difference = \frac{100 * (fluorescence_{sample} - fluorescence_{untreated\ control})}{fluorescence_{untreated\ control}}$$

**Cytoplasmic membrane depolarization assays**

The cytoplasmic membrane depolarization assay was performed using the membrane potential-sensitive dye 3,3'-dipropylthiadicarbocyanine iodide (DiSC$_3$-5). *A. baumannii* ATCC 19606 and *P. aeruginosa* PAO1 in the mid-logarithmic phase were washed and resuspended at 0.05 OD mL$^{-1}$ (optical value at 600 nm) in HEPES buffer (pH 7.2) containing 20 mmol L$^{-1}$ glucose and 0.1 mol L$^{-1}$ KCl. DiSC$_3$-5 at 20 μmol L$^{-1}$ was added to the bacterial suspension (100 μL per well) for 15 min to stabilize the fluorescence which indicates the incorporation of the dye into the bacterial membrane, and then the peptides were mixed 1:1 with the bacteria to a final concentration corresponding to their MIC$_{100}$ values. Membrane depolarization was then followed by reading changes in the fluorescence ($\lambda_{ex}$ = 622 nm, $\lambda_{em}$ = 670 nm) over time for 60 min. The relative fluorescence was calculated using the untreated control (buffer + bacteria + fluorescent dye) as baseline and the following equation was applied to reflect % of difference between the baselines and the sample:

$$Percentage\ difference = \frac{100 * (fluorescence_{sample} - fluorescence_{untreated\ control})}{fluorescence_{untreated\ control}}$$

**Data availability**

This study did not generate new unique reagents. The data used in this study are available from two main sources. The raw data were collected from the following open-source antimicrobial peptide (AMP) databases: DRAMP (http://dramp.cpu-bioinfor.org/), DADP (http://bio.ynu.edu.cn/dadp), LAMP2 (http://biotechlab.fudan.edu.cn/database/lamp/), dbAMPv2.0 (https://dbamp.cpu-bioinfor.org/), and AMPScanner (https://www.dveltri.com/ascan/v2/ascan.html).. Further information and requests for resources should be directed to the lead contact, Cesar de la Fuente-Nunez (cfuente@upenn.edu).

**Code availability**

ApexAmphion is available at GitLab (https://gitlab.com/chq1155/AMPGen_Product.git).

**Acknowledgments**

C.F.-N. holds a Presidential Professorship at the University of Pennsylvania. Research reported in this publication was supported by the NIH R35GM138201 and DTRA HDTRA1-21-1-0014. We thank de la Fuente Lab members for insightful discussions. Figures created with BioRender.com are attributed as such. The work described in this paper was supported in part by the Research Grants Council of the Hong Kong Special Administrative Region, China, under Project T45-401/22-N.

**Author contributions**

H.C., J. Z., Z. G., F. W., C. B., G. C., J. L., Y. C., and P. H. designed the method. H. C., J. Z., Z. G. wrote the program. M.D.T.T. and C.F.-N. designed the experimental validation. M.D.T.T. performed experiments and interpreted the data. H.C., M.D.T.T., and C.F.-N. wrote the manuscript and all other authors revised it.

**Competing interests**

C.F.-N. is a co-founder and scientific advisor to Peptaris, Inc., provides consulting services to Invaio Sciences and is a member of the Scientific Advisory Boards of Nowture S.L., Peptidus, and Phare Bio. C.F.-N. is also on the Advisory Board of the Peptide Drug Hunting Consortium (PDHC). M.D.T.T. is a co-founder and scientific advisor to Peptaris, Inc. All other authors declare no competing interests.

**References**

1. O'Neill, J. Tackling drug-resistant infections globally: final report and recommendations. *Rev. Antimicrob. Resist.* (2016).
2. World Health Organization and Food and Agriculture Organization of the United Nations and World Organisation for Animal Health. *Technical Series on Antimicrobial Resistance: The Impact of Antimicrobial Resistance on Global Health and Food Security.* (2019).
3. de la Fuente-Nunez, C. & Collins, J. J. Essay: Using Machine Learning for Antibiotic Discovery. *Phys. Rev. Lett.* **135**, 030001 (2025).
4. Wan, F., Torres, M. D. T., Peng, J. & de la Fuente-Nunez, C. Deep-learning-enabled antibiotic discovery through molecular de-extinction. *Nat. Biomed. Eng.* **8**, 854–871 (2024).
5. Wong, F., de la Fuente-Nunez, C. & Collins, J. J. Leveraging artificial intelligence in the fight against infectious diseases. *Science* **381**, 164–170 (2023).
6. Porto, W. F. *et al.* In silico optimization of a guava antimicrobial peptide enables combinatorial exploration for peptide design. *Nat. Commun.* **9**, 1490 (2018).
7. Mahlapuu, M., Håkansson, J., Ringstad, L. & Bj"orn, C. Antimicrobial peptides: an emerging category of therapeutic agents. *Front. Cell. Infect. Microbiol.* **6**, 194 (2016).
8. Huang, J. *et al.* Identification of potent antimicrobial peptides via a machine-learning pipeline that mines the entire space of peptide sequences. *Nat. Biomed. Eng.* 1–14 (2023).

9. Wang, B. *et al.* Explainable deep learning and virtual evolution identifies antimicrobial peptides with activity against multidrug-resistant human pathogens. *Nat. Microbiol.* 1–16 (2025).
10. Olayo-Alarcon, R. *et al.* Pre-trained molecular representations enable antimicrobial discovery. *Nat. Commun.* **16**, 3420 (2025).
11. Santos-Junior, C. D., Pan, S., Zhao, X.-M. & Coelho, L. P. Macrel: antimicrobial peptide screening in genomes and metagenomes. *PeerJ* **8**, e10555 (2020).
12. Santos-Júnior, C. D. *et al.* Discovery of antimicrobial peptides in the global microbiome with machine learning. *Cell* **187**, 3761-3778.e16 (2024).
13. Torres, M. D. T. *et al.* Mining human microbiomes reveals an untapped source of peptide antibiotics. *Cell* **187**, 5453-5467.e15 (2024).
14. Durrant, M. G. & Bhatt, A. S. Automated prediction and annotation of small open reading frames in microbial genomes. *Cell Host Microbe* **29**, 121–131 (2021).
15. Lawrence, T. J. *et al.* amPEPpy 1.0: a portable and accurate antimicrobial peptide prediction tool. *Bioinformatics* **37**, 2058–2060 (2021).
16. Cesaro, A., Wan, F., Torres, M. D. & de la Fuente-Nunez, C. Design of multimodal antibiotics against intracellular infections using deep learning. *bioRxiv* 2024–12 (2024).
17. Torres, M. D. T., Wan, F. & de la Fuente-Nunez, C. Deep learning reveals antibiotics in the archaeal proteome. *Nat. Microbiol.* **10**, 2153–2167 (2025).
18. Guan, C., Torres, M. D. T., Li, S. & de la Fuente-Nunez, C. Computational exploration of global venoms for antimicrobial discovery with Venomics artificial intelligence. *Nat. Commun.* **16**, 6446 (2025).
19. Fjell, C. D., Jenssen, H., Cheung, W. A., Hancock, R. E. & Cherkasov, A. Optimization of antibacterial peptides by genetic algorithms and cheminformatics. *Chem. Biol. Drug Des.* **77**, 48–56 (2011).
20. Boone, K., Wisdom, C., Camarda, K., Spencer, P. & Tamerler, C. Combining genetic algorithm with machine learning strategies for designing potent antimicrobial peptides. *BMC Bioinformatics* **22**, 239 (2021).
21. Szymczak, P. *et al.* Discovering highly potent antimicrobial peptides with deep generative model HydrAMP. *Nat. Commun.* **14**, 1453 (2023).
22. Van Oort, C. M., Ferrell, J. B., Remington, J. M., Wshah, S. & Li, J. AMPGAN v2: Machine Learning-Guided Design of Antimicrobial Peptides. *J. Chem. Inf. Model.* **61**, 2198–2207 (2021).
23. Das, P. *et al.* PepCVAE: Semi-Supervised Targeted Design of Antimicrobial Peptide Sequences. Preprint at http://arxiv.org/abs/1810.07743 (2018).
24. Wang, R. *et al.* Diff-AMP: tailored designed antimicrobial peptide framework with all-in-one generation, identification, prediction and optimization. *Brief. Bioinform.* **25**, bbae078 (2024).
25. Li, T. *et al.* A foundation model identifies broad-Spectrum antimicrobial peptides against drug-resistant bacterial infection. *Nat. Commun.* **15**, 7538 (2024).
26. Wang, J. *et al.* Discovery of antimicrobial peptides with notable antibacterial potency by an LLM-based foundation model. *Sci. Adv.* **11**, eads8932 (2025).
27. Chen, T., Vure, P., Pulugurta, R. & Chatterjee, P. AMP-diffusion: Integrating latent diffusion with protein language models for antimicrobial peptide generation. *bioRxiv* 2024–03 (2024).
28. Torres, M. D. T. *et al.* A generative artificial intelligence approach for antibiotic optimization. Preprint at https://doi.org/10.1101/2024.11.27.625757 (2024).

29. Pirtskhalava, M. *et al.* DBAASP v3: database of antimicrobial/cytotoxic activity and structure of peptides as a resource for development of new therapeutics. *Nucleic Acids Res.* **49**, D288–D297 (2021).
30. Shi, G. *et al.* DRAMP 3.0: an enhanced comprehensive data repository of antimicrobial peptides. *Nucleic Acids Res.* **50**, D488–D496 (2022).
31. Wang, G., Li, X. & Wang, Z. APD3: the antimicrobial peptide database as a tool for research and education. *Nucleic Acids Res.* **44**, D1087–D1093 (2016).
32. Thomas, S., Karnik, S., Barai, R. S., Jayaraman, V. K. & Idicula-Thomas, S. CAMP: a useful resource for research on antimicrobial peptides. *Nucleic Acids Res.* **38**, D774–D780 (2010).
33. Peng, S. & Rajjou, L. Unifying antimicrobial peptide datasets for robust deep learning-based classification. *Data Brief* **56**, 110822 (2024).
34. Xiao, B. *et al.* A comprehensive dataset of therapeutic peptides on multi-function property and structure information. *Sci. Data* **12**, 1213 (2025).
35. Nijkamp, E., Ruffolo, J. A., Weinstein, E. N., Naik, N. & Madani, A. ProGen2: exploring the boundaries of protein language models. *Cell Syst.* **14**, 968–978 (2023).
36. Ouyang, L. *et al.* Training language models to follow instructions with human feedback. *Adv. Neural Inf. Process. Syst.* **35**, 27730–27744 (2022).
37. Krishnan, A. *et al.* A generative deep learning approach to de novo antibiotic design. *Cell* S0092867425008554 (2025) doi:10.1016/j.cell.2025.07.033.
38. Gawde, U. *et al.* CAMPR4: a database of natural and synthetic antimicrobial peptides. *Nucleic Acids Res.* **51**, D377–D383 (2023).
39. Veltri, D., Kamath, U. & Shehu, A. Deep learning improves antimicrobial peptide recognition. *Bioinformatics* **34**, 2740–2747 (2018).
40. Müller, A. T., Gabernet, G., Hiss, J. A. & Schneider, G. modlAMP: Python for antimicrobial peptides. *Bioinformatics* **33**, 2753–2755 (2017).
41. Suzek, B. E., Huang, H., McGarvey, P., Mazumder, R. & Wu, C. H. UniRef: comprehensive and non-redundant UniProt reference clusters. *Bioinformatics* **23**, 1282–1288 (2007).
42. Steinegger, M. & Söding, J. MMseqs2 enables sensitive protein sequence searching for the analysis of massive data sets. *Nat. Biotechnol.* **35**, 1026–1028 (2017).
43. Novković, M., Simunić, J., Bojović, V., Tossi, A. & Juretić, D. DADP: the database of anuran defense peptides. *Bioinformatics* **28**, 1406–1407 (2012).
44. Ye, G. *et al.* LAMP2: a major update of the database linking antimicrobial peptides. *Database* **2020**, baaa061 (2020).
45. Jhong, J.-H. *et al.* dbAMP 2.0: updated resource for antimicrobial peptides with an enhanced scanning method for genomic and proteomic data. *Nucleic Acids Res.* **50**, D460–D470 (2022).

## Tables

**Table 1. Similarity test of ApexAmphion's candidates.** All hits are shown with Query, Target, Identity percentage, Length, E-value, and Bits.

| Query | Target | %Identity | Length | E-value | Bits |
|---|---|---|---|---|---|
| amphionin-1 | UniRef50_P86170 | 94.444 | 18 | 0.044 | 34.3 |
| amphionin-6 | UniRef50_Q8UUG0 | 100 | 22 | 3.54E-07 | 48.5 |
| amphionin-6 | UniRef50_A0A8C4DW39 | 95.455 | 22 | 1.17E-05 | 45.4 |
| amphionin-6 | UniRef50_UPI0037044B02 | 80 | 20 | 0.49 | 33.9 |
| amphionin-6 | UniRef50_A0AAD3MA38 | 65.385 | 26 | 0.65 | 33.9 |
| amphionin-6 | UniRef50_A0AAD6AEG7 | 54.545 | 22 | 1 | 32.3 |
| amphionin-6 | UniRef50_A0A4W6DEI7 | 65.385 | 26 | 1.6 | 31.6 |
| amphionin-6 | UniRef50_P0DUJ5 | 59.091 | 22 | 7.4 | 29.6 |
| amphionin-6 | UniRef50_A0A267H675 | 55.556 | 18 | 9.9 | 30.4 |
| amphionin-11 | UniRef50_Q5SC60 | 100 | 21 | 1.25E-05 | 46.6 |
| amphionin-12 | UniRef50_Q5SC60 | 100 | 19 | 7.02E-04 | 42 |
| amphionin-18 | UniRef50_C0HK42 | 93.333 | 15 | 6.7 | 28.9 |
| amphionin-85 | UniRef50_A0A8J5J858 | 75 | 12 | 4.9 | 30.8 |
| amphionin-94 | UniRef50_UPI001B8604BB | 81.25 | 16 | 0.7 | 33.1 |

# Figures

## a

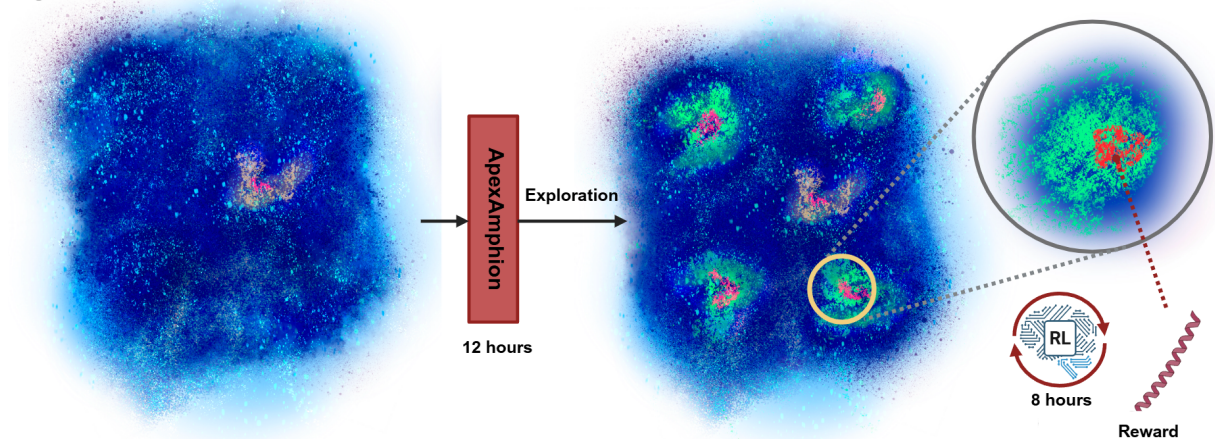

## b

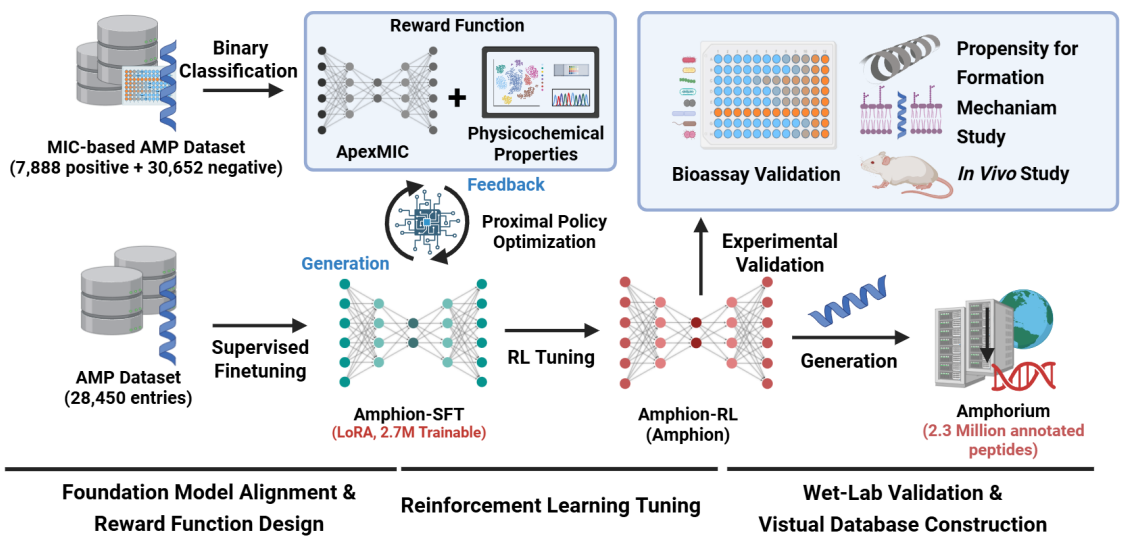

**Figure 1. Overview of ApexAmphion: a)** ApexAmphion leverages the limited scale of known AMPs (Yellow) to explore the hidden AMPs (Green) in protein universe and the low-MIC AMP subset (Red) based on pretrained large protein language models (ProGen2-XL) and reinforcement learning technique, respectively. The first stage involves supervised fine-tuning based on the known AMPs, taking 12 hours to train. The second stage involves Poximal Policy Optimization (PPO) to deeply explore the low-MIC candidates in the shed-lighted AMP distributions, taking 8 hours to for tuning. **b)** ApexAmphion applies a three-stage scheme to leverage the computational power of large protein language models. In the foundation model alignment stage, AMP sequences are applied to tune the base model. In the reward function design stage, ApexMIC is trained based on ESM2-8M to conduct binary classification. In the RL tuning stage, Amphion-SFT is tuned by PPO algorithm, using the reward function composed of ApexMIC and physicochemical properties. Then, the generated samples from Ampion-RL is screened and ranked by ApexMIC for experimental validation. And the generated samples from Amphion-SFT and Amphion-RL are curated and constructed into a 2.3 million-scale virtual peptide database--Amphorium. Fugure was created with BioRender.com.

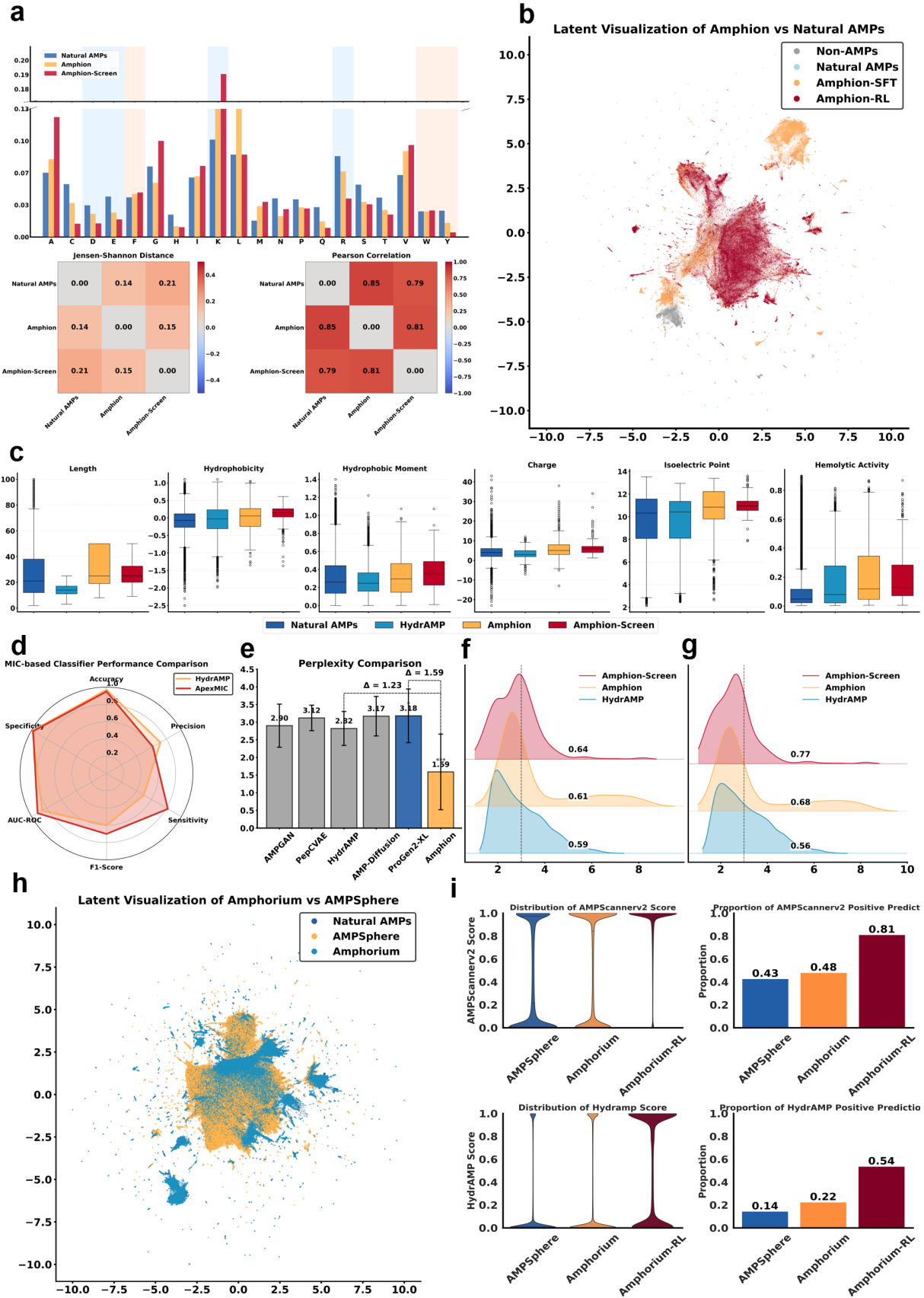

**Figure 2. Main computational experiments on ApexAmpion platform. a)** The amino acid frequency distribution between Amphion's samples and the natural AMPs. **b)** UMAP visualization of Amphion-SFT, Amphion-RL against natural AMPs and Non-AMPs under ESM-8M's representation. **c)** The property distribution of Amphion's samples and the natural AMPs. **d)** Performance comparison of ApexMIC to HydrAMP's classifier on low-MIC binary identification task. **e)** Benchmark for inference cross-entropy among Amphion, ProGen2-XL and other AMP generation baselines. **f-g)** The latent distance distribution between Amphion, Amphion-Screen (The screened candidates for wet-lab experiments), and HydrAMP against natural AMPs and natural low MIC AMPs under ESM2-8M's representation. **h)** UMAP visualization of Amphion's samples, AMPSphere, and natural AMPs. **i)** Predicted distribution of Amphorium, Amphorium-RL, and AMPSphere under AMPScannerv2 (Binary AMP prediction) and HydrAMP's classifier (Binary Low-MIC AMP prediction).

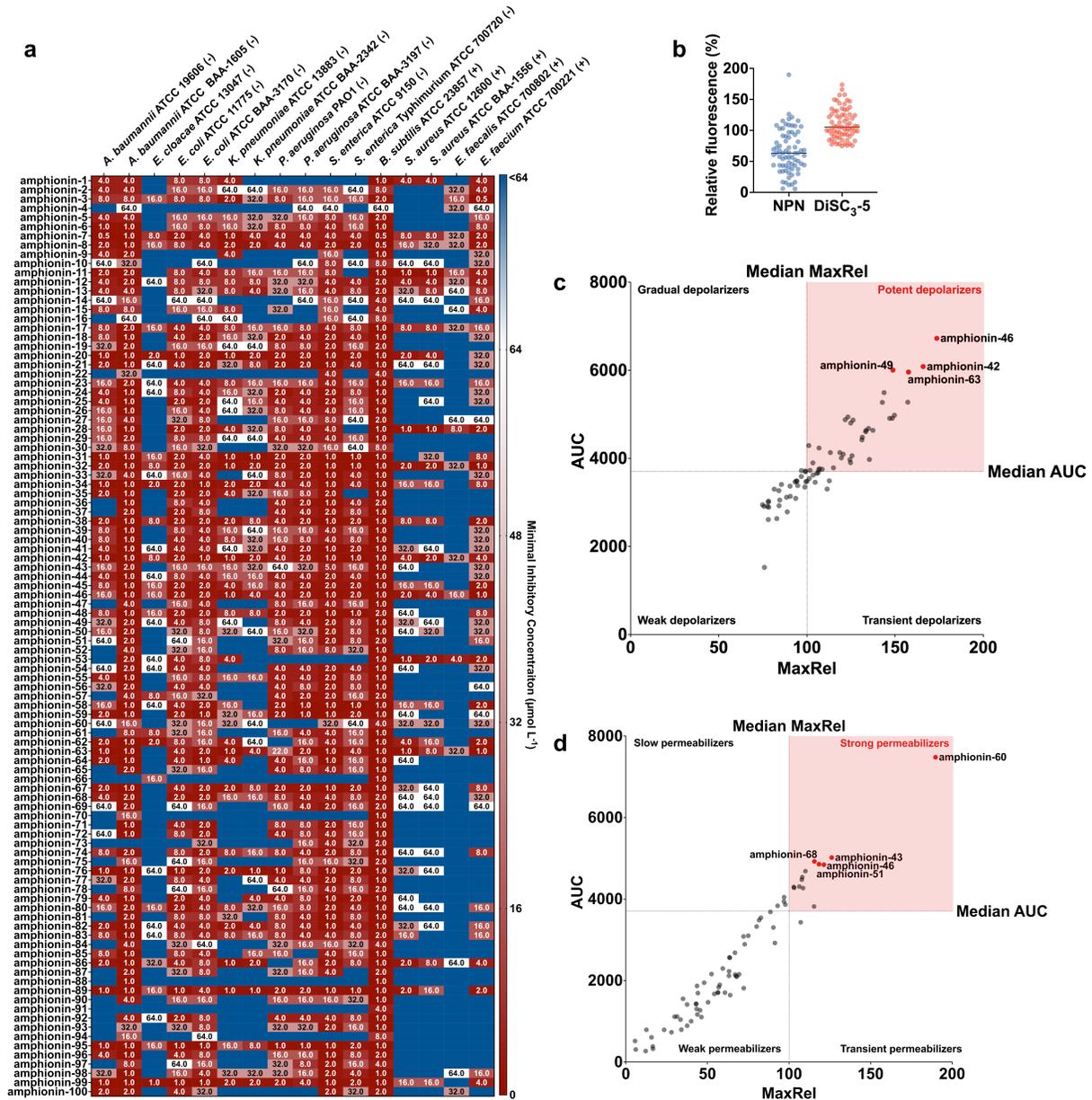

**Figure 3. Antimicrobial activity and membrane-disruptive effect of amphionins. a)** Heat map showing the antimicrobial activities (µmol L$^{-1}$) of active amphionins against 16 clinically relevant pathogens, including Gram-negative (indicated by –) and Gram-positive (indicated by +) susceptible and antibiotic-resistant strains. Briefly, 10$^5$ bacterial cells were incubated with serially diluted peptides (0-64 µmol L$^{-1}$) at 37 °C. Bacterial growth was assessed by measuring the optical density at 600 nm in a microplate reader at 1 day post-treatment. The MIC values presented in the heat map represent the mode of the replicates for each condition. **b)** To assess whether amphionins act on bacterial membranes, all active peptides against *A. baumannii* ATCC 19606 were subjected to outer membrane permeabilization and cytoplasmic membrane depolarization assays. Amphionins showed higher depolarization compared to permeabilization effects. **c)** Scatter plot of cytoplasmic membrane depolarization (DiSC$_3$-5 assay). Each point represents an amphionin, with MaxRel (maximum relative fluorescence) plotted against AUC (integrated fluorescence). Dashed lines indicate median values used to divide peptides into four mechanistic categories: potent depolarizers (high peak and sustained disruption), transient depolarizers (strong but short-lived), gradual depolarizers (steady but moderate), and weak depolarizers (minimal effect). Representative amphionins from the potent group, i.e., strong depolarizers are labeled. **d)** Scatter plot of outer-membrane permeabilization (NPN uptake), analyzed as in **c**. Quadrants define strong permeabilizers (robust and sustained outer membrane damage), transient permeabilizers (sharp but short-lived response), slow permeabilizers (gradual accumulation without a strong peak), and weak permeabilizers (little or no activity). Representative amphionins are labeled.

# Supplementary Information

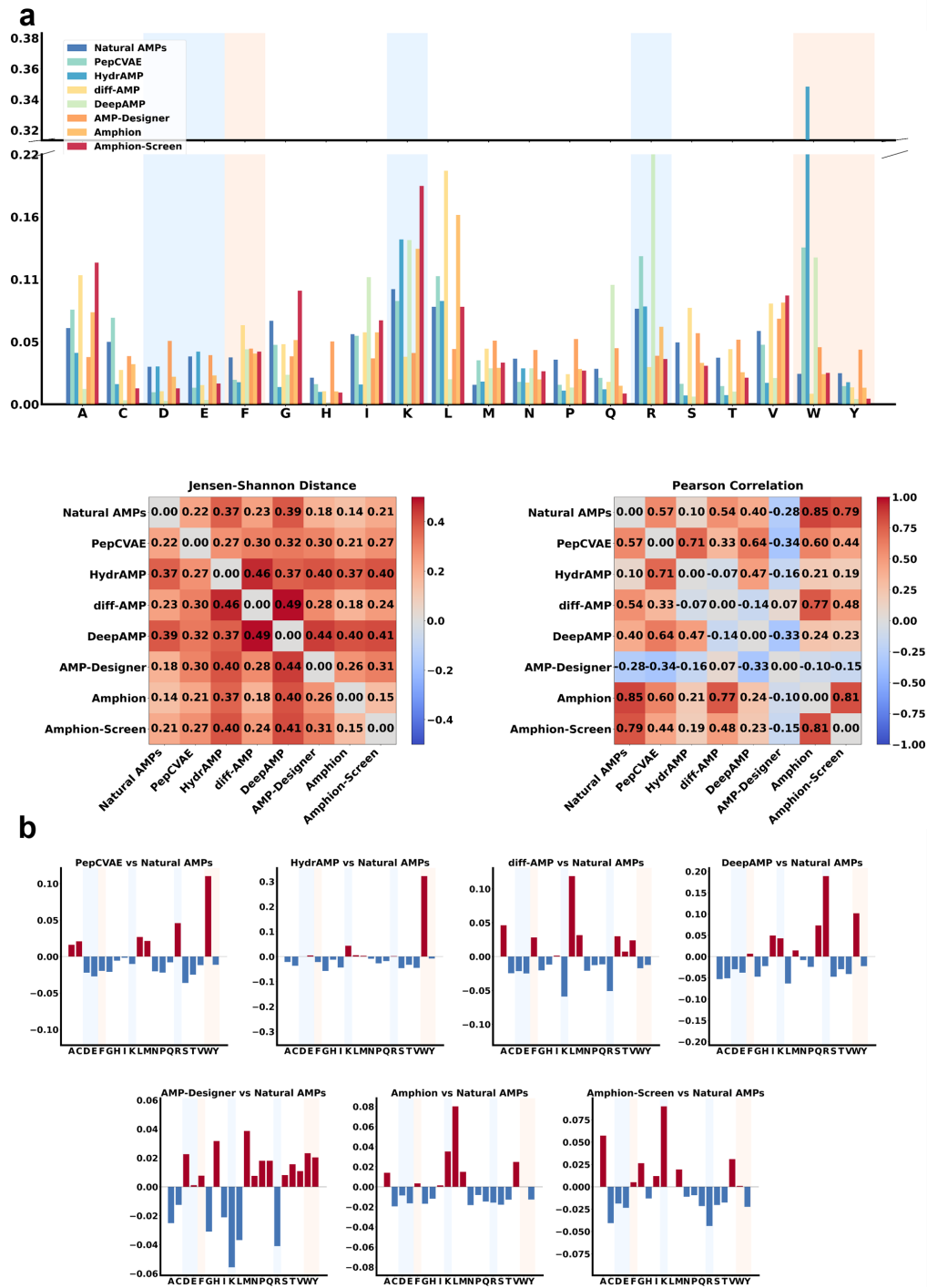

**Supplementary Figure 1. Supplementary experiments on ApexAmphion's Generation. a)** The amino acid frequency distribution between Amphion's and other baselines' samples and the natural AMPs. **b)** The difference of amino acid frequency distribution between Amphion's and other baselines' samples and the natural AMPs.

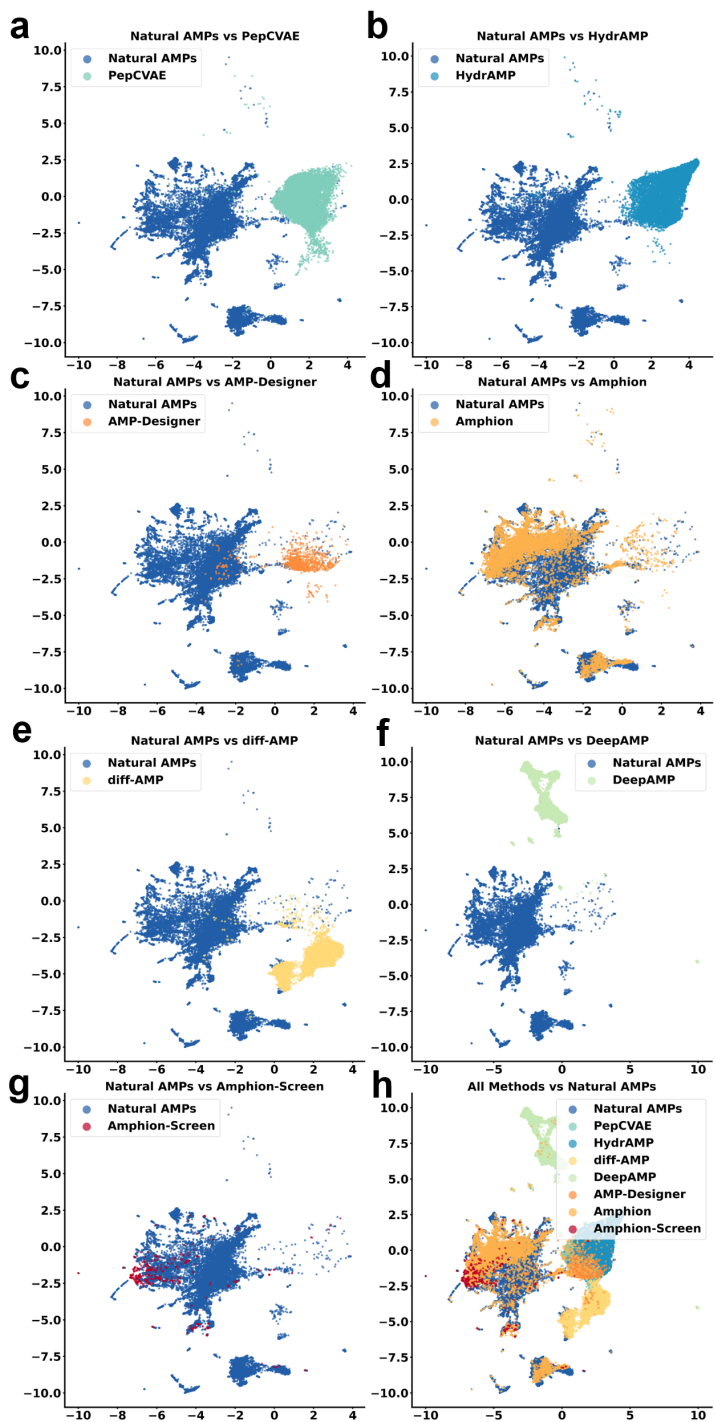

**Supplementary Figure 2. Supplementary UMAP visualization of generated samples of AMP generation methods against natural AMPs. a-g)** Visualization of PepCVAE, HydrAMP, AMP-Designer, diff-AMP, DeepAMP, Amphion, Amphion-Screen against natural AMPs. **h)** Combined plot of all methods and natural AMPs.

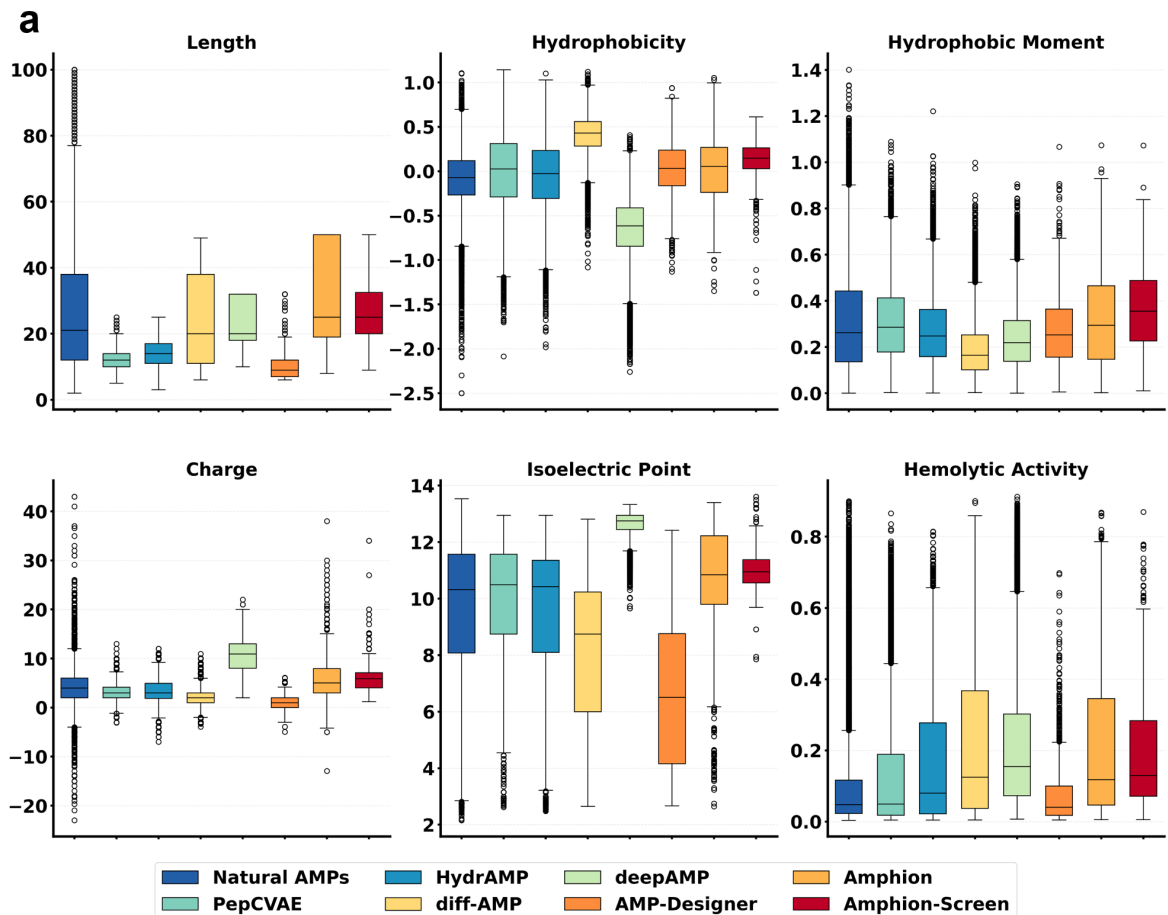
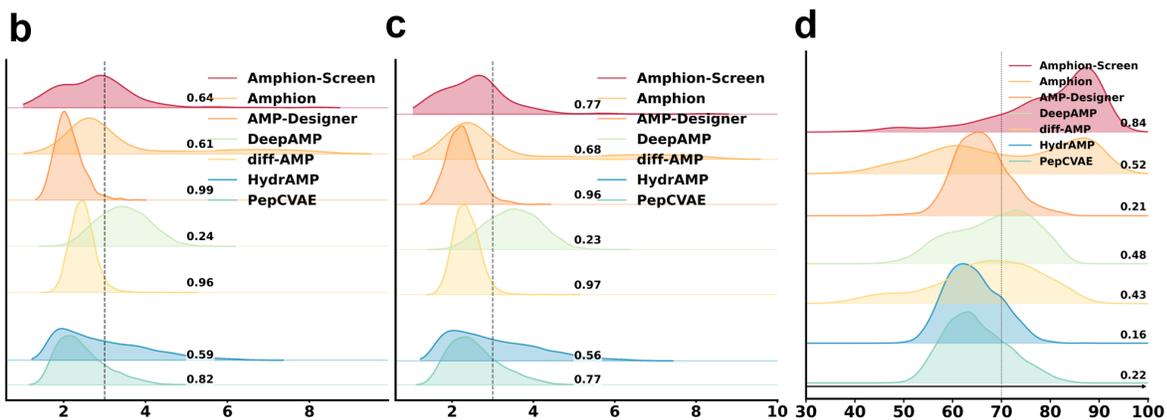

**Supplementary Figure 3. Supplementary experiments on ApexAmphion's generation. a)** The property distribution (Length, Hydrophobicity, Hydrophobic moment, Net charge, Isoletric point, and Hemolytic activity) for all computational AMP generative methods. **b-d)** The latent distribution visualization under Amphion's samples against the other computational generative baselines according to the latent distance to Natural AMPs, the latent distance to natural low-MIC AMPs, and the pLDDT scores. The number on the right of each distribution denote the proportion under (B-C) and surpass (D) the thresholds.

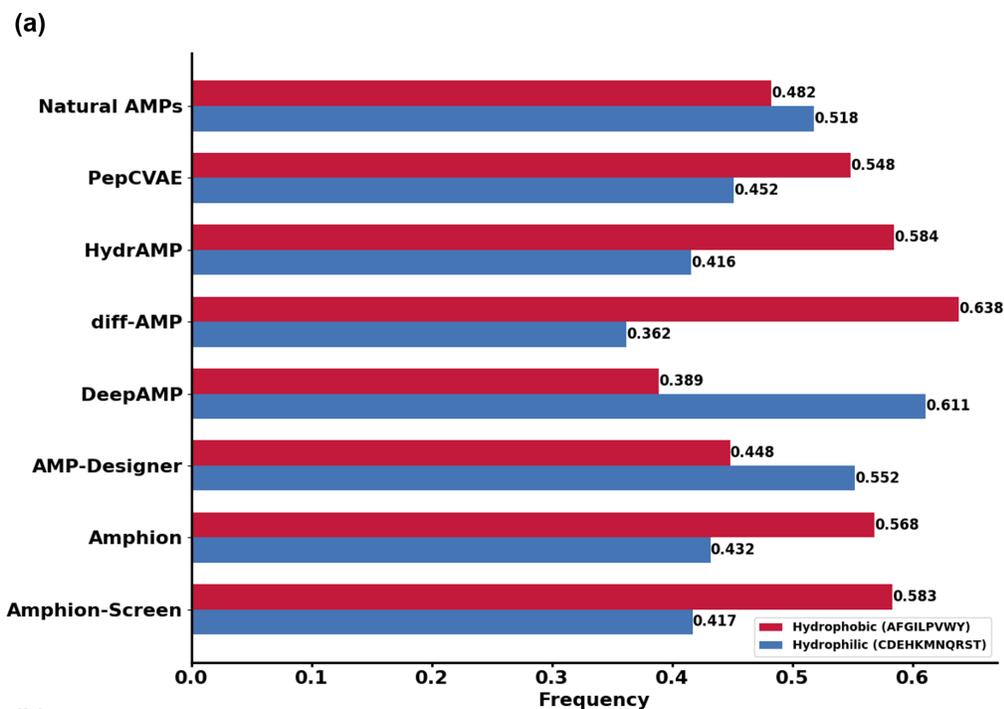

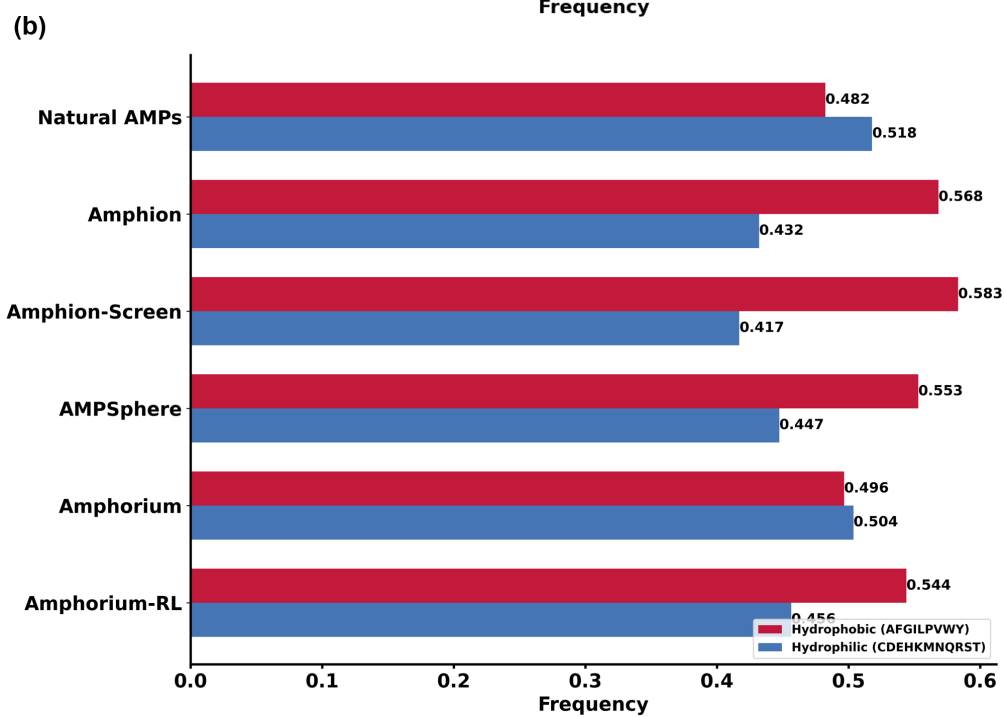

**Figure 4. Supplementary experiments on hydrophobic and hydrophilic amino acid frequency. a)** The hydrophobic and hydrophilic amino acid frequency distribution among Amphion, Amphion-Screen, and other AMP generation methods against the natural AMPs. **b)** The hydrophobic and hydrophilic amino acid frequency distribution between Amphorium, Amphorium-RL, Amphion, Amphion-Screen, and AMPSphere against the natural AMPs.

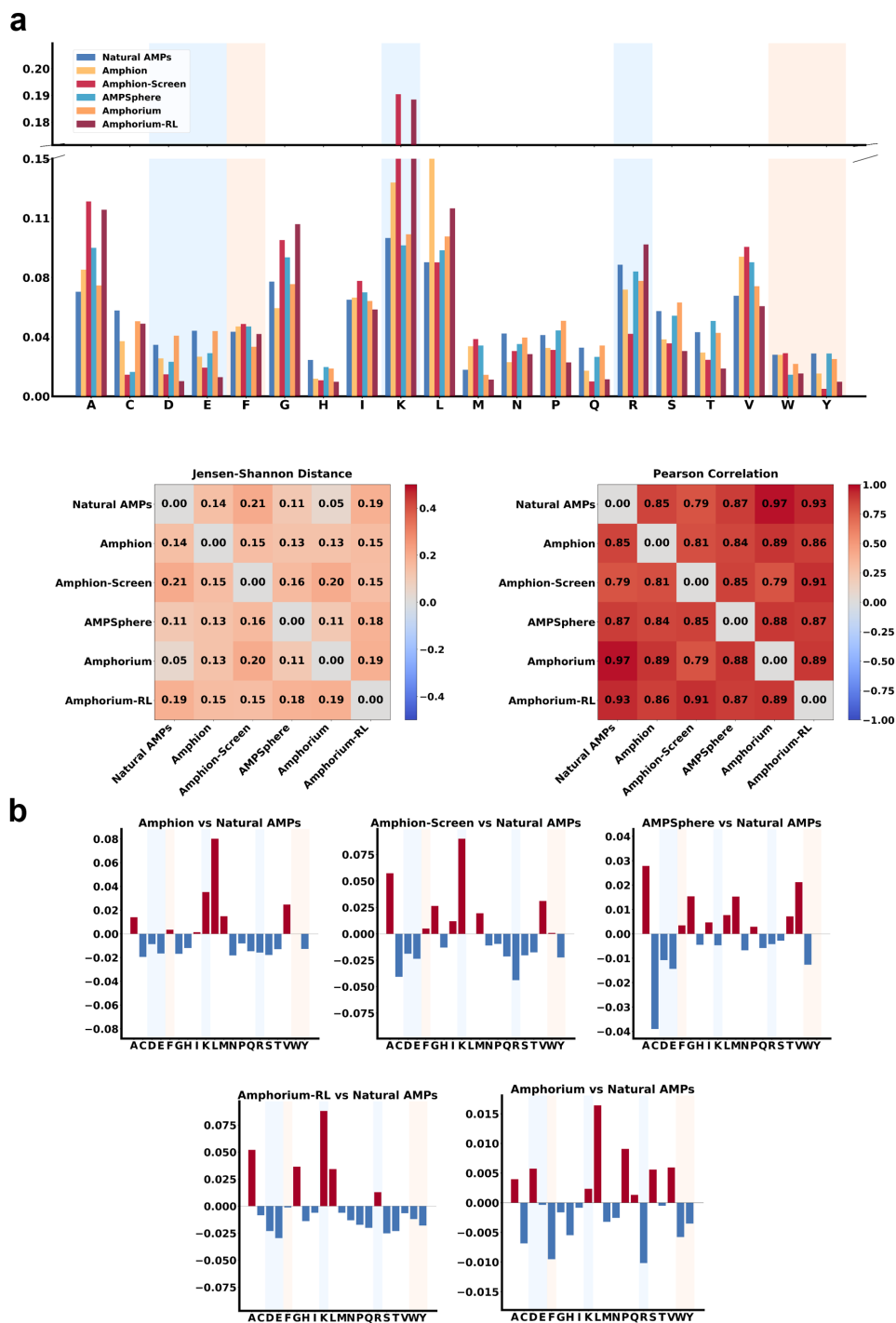

**Supplementary Figure 5. Supplementary experiments on ApexAmphorium's entries. a)** The amino acid frequency distribution between Amphorium, Amphorium-RL, Amphion, Amphion-Screen, and AMPSphere against the natural AMPs. **(b)** The difference of amino acid frequency distribution between Amphorium, Amphorium-RL, Amphion, Amphion-Screen, and AMPSphere against the natural AMPs.

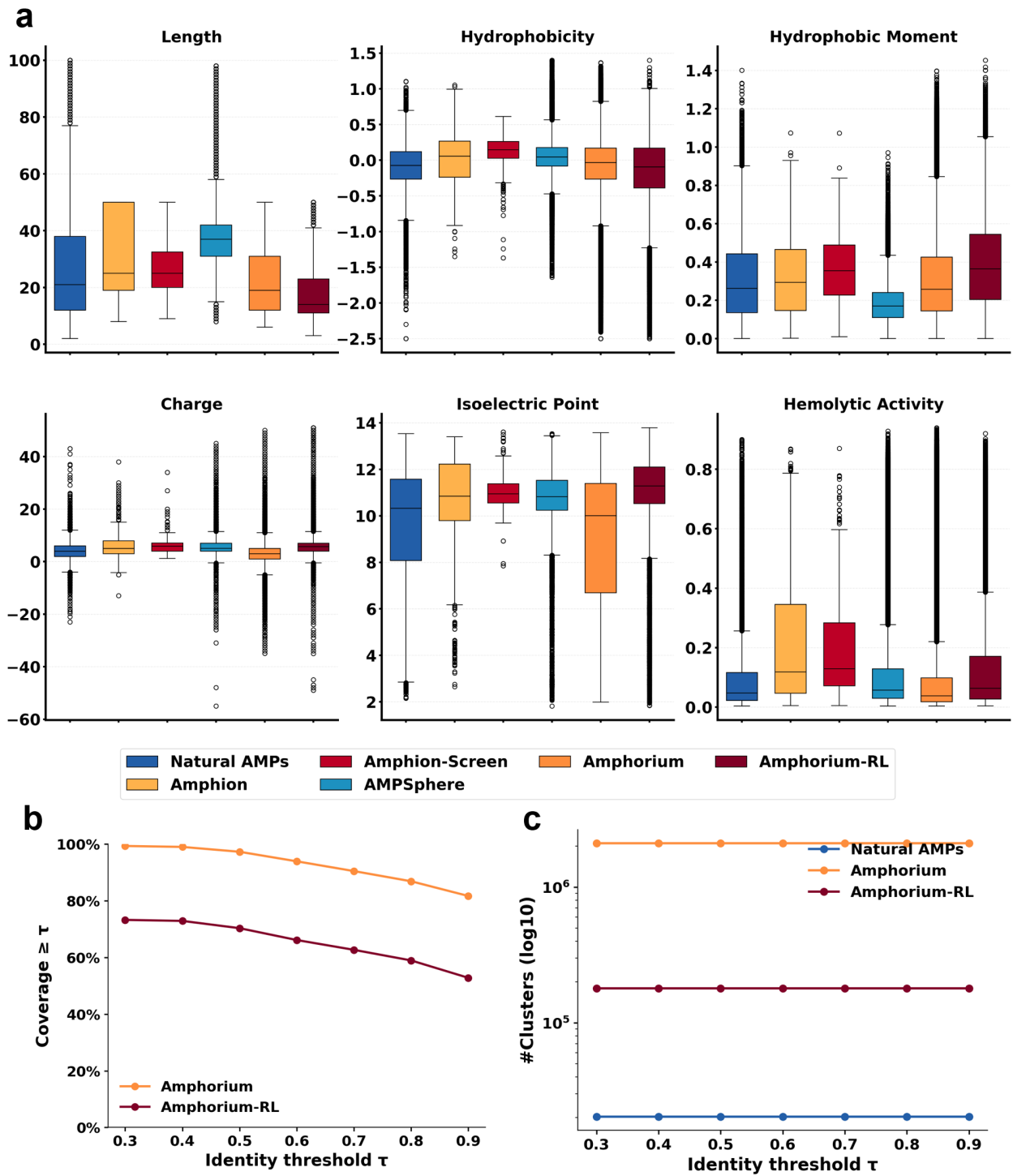

**Supplementary Figure 6. Supplementary experiments on ApexAmphorium's entries. a)** The property distribution (including length, hydrophobicity, hydrophobic moment, charge, isoeletric point, and hemolytic activity) between Amphorium, Amphorium-RL, Amphion, Amphion-Screen, and AMPSphere against the natural AMPs. **b-c)** Novelty and diversity analysis of Amphorium against natural AMPs by MMseqs2.

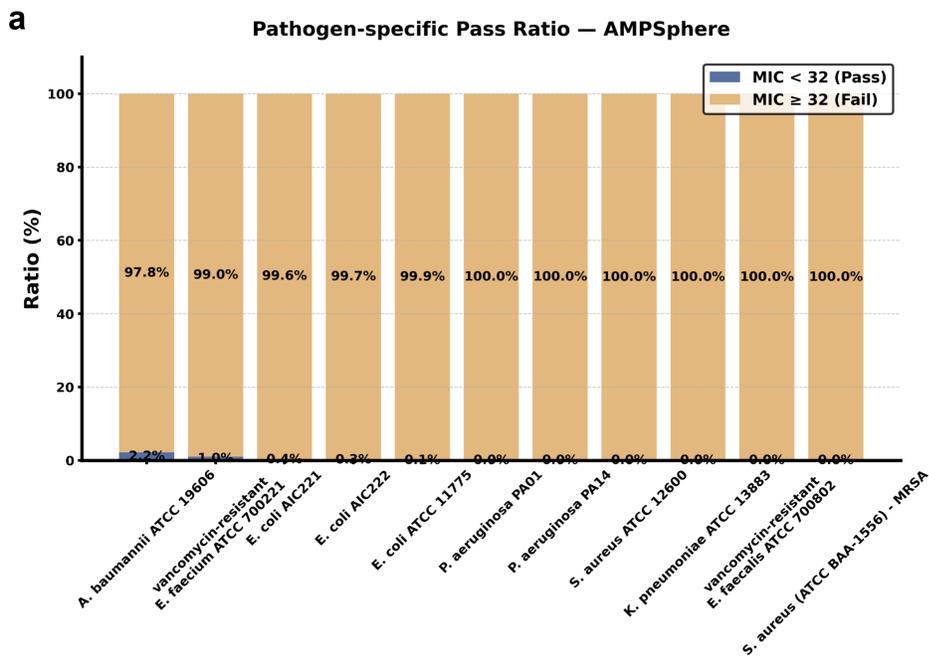

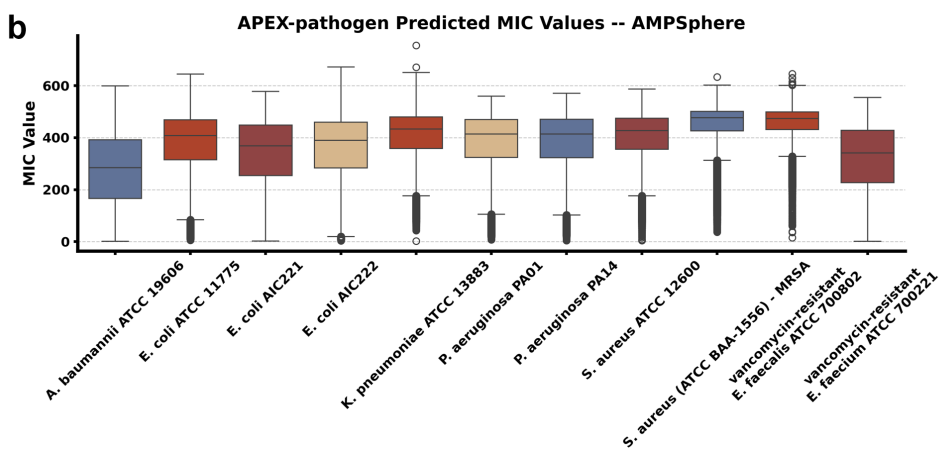

**Supplementary Figure 7. Supplementary results of Apex 1.1 annotated distribution on AMPSphere entries. a)** The pass ratio of AMPSphere entries on each pathogens under the condition of MIC <32 μmol L$^{-1}$. **b)** The predicted MIC value distribution of AMPSphere entries on each pathogens.

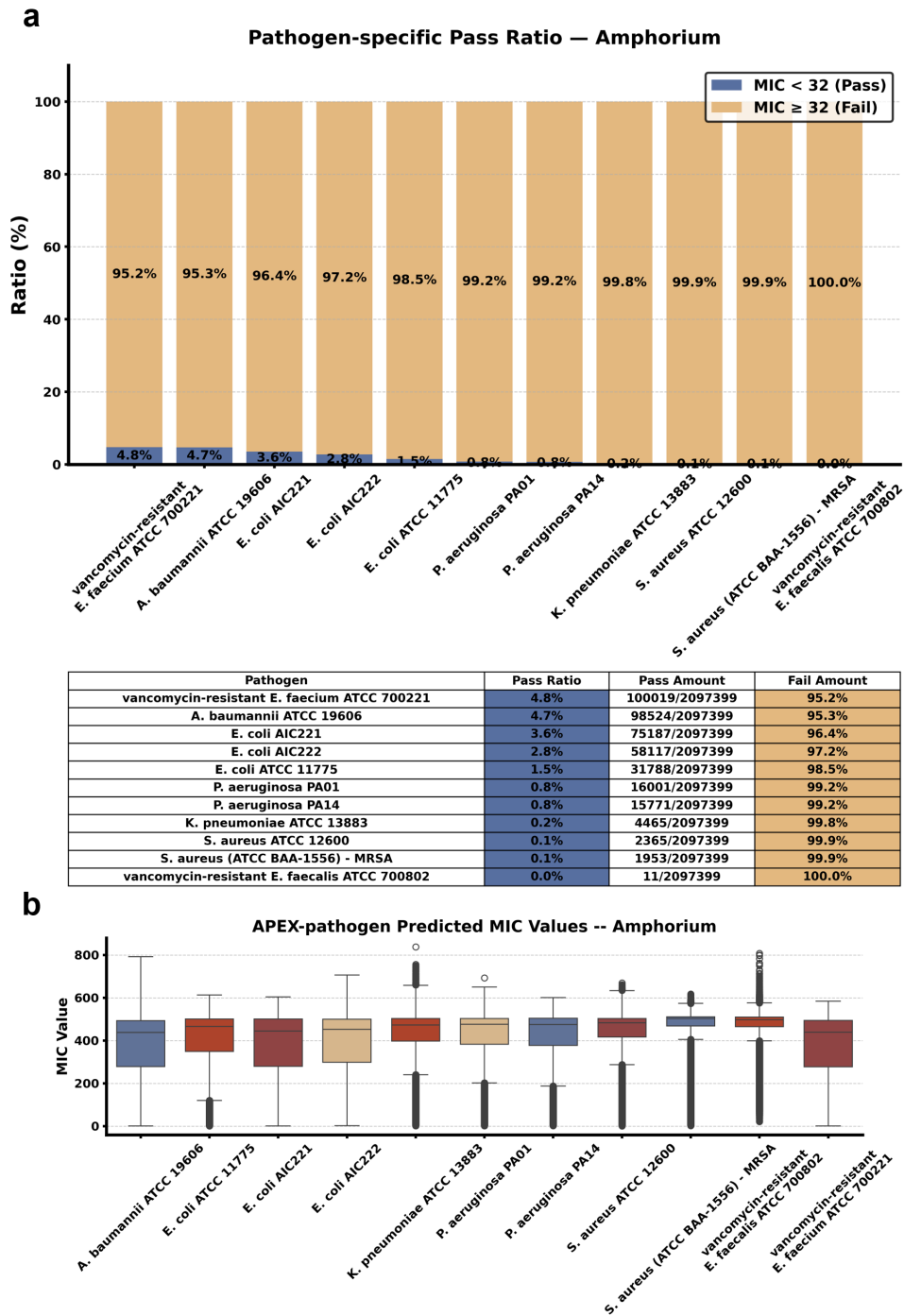

**Supplementary Figure 8. Supplementary results of Apex 1.1 annotated distribution on ApexAmphorium entries. a)** The pass ratio of ApexAmphorium entries on each pathogens under the condition of MIC <32 μmol L$^{-1}$. **b)** The predicted MIC value distribution of ApexAmphorium entries on each pathogens.

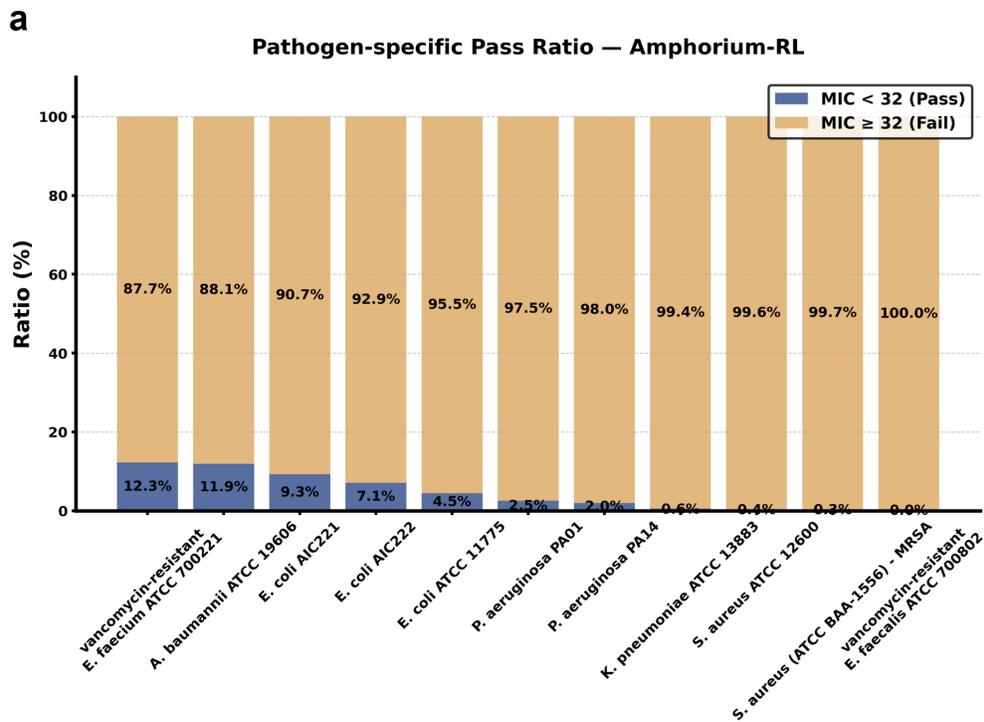

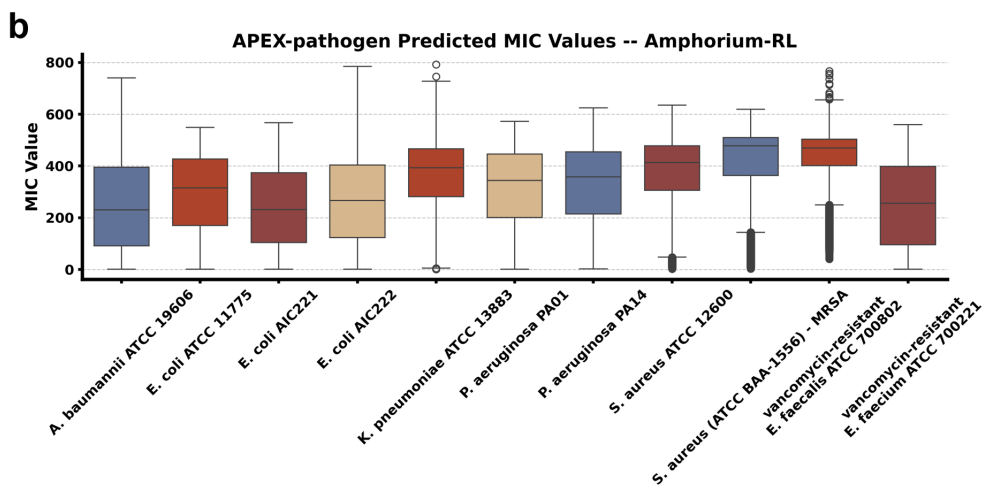

**Supplementary Figure 9. Results of Apex 1.1 annotated distribution on ApexAmphorium-RL's entries. a)** The pass ratio of ApexAmphorium-RL entries on each pathogens under the condition of MIC <32 μmol L$^{-1}$. **b)** The predicted MIC value distribution of ApexAmphorium-RL entries on each pathogens.

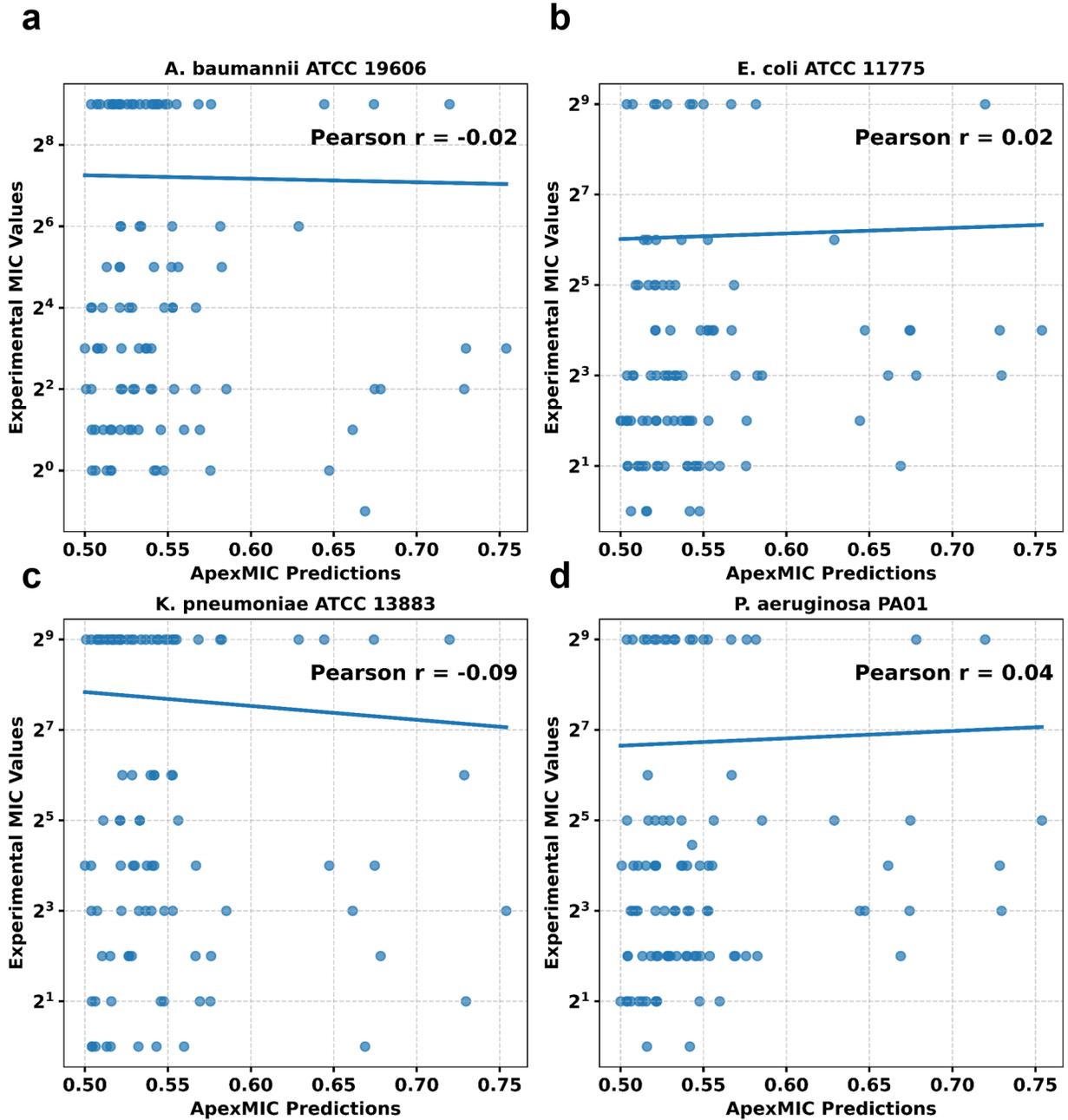

**Supplementary Figure 10. Correlation analysis between ApexMIC's prediction and wet-lab experimental MIC values of amphionins. a-d)** Sub-correlation between ApexMIC's predicted scores and tested MICs on different pathogens.

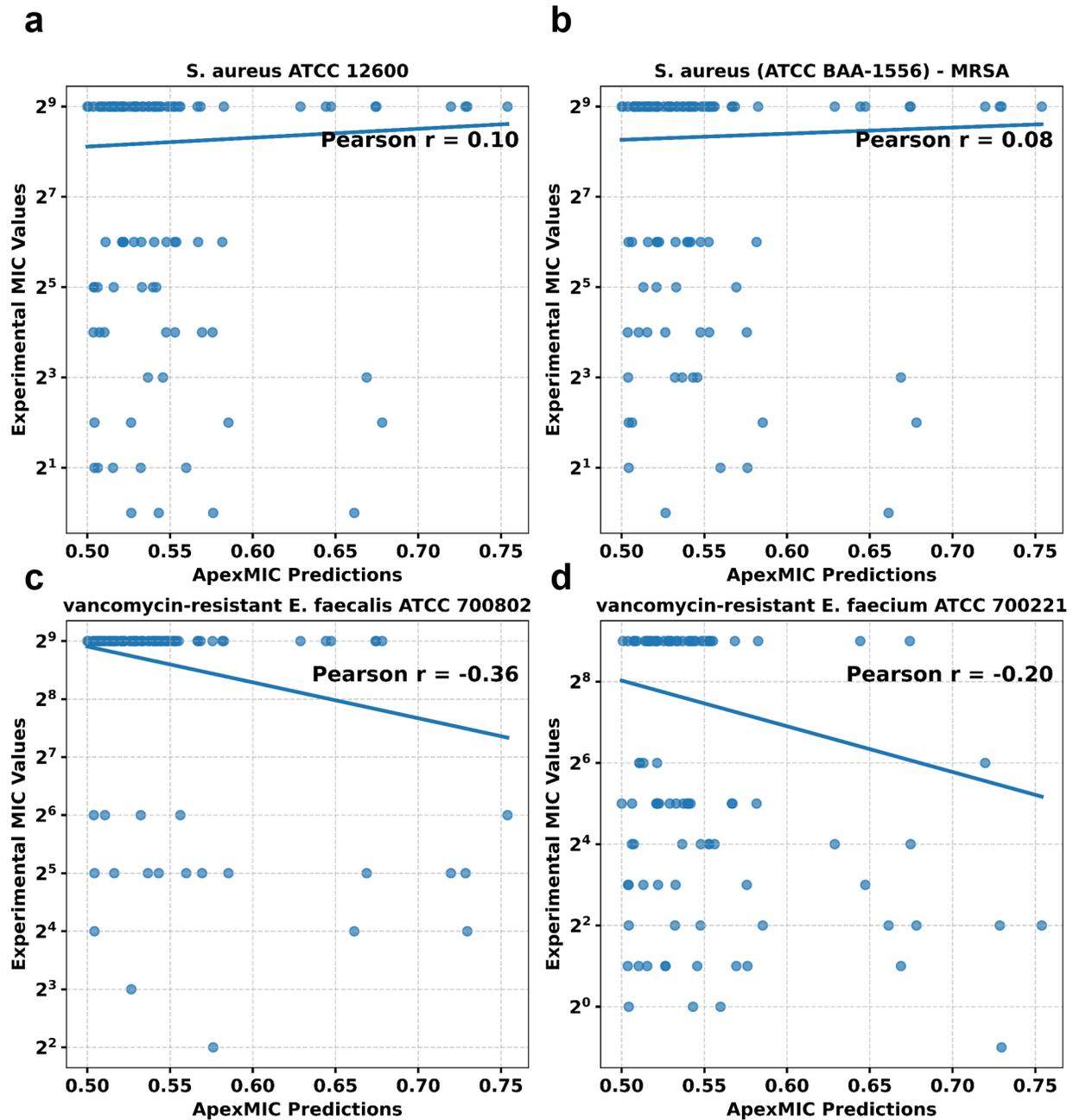

**Supplementary Figure 11. Correlation analysis between ApexMIC's prediction and wet-lab experimental MIC values of amphionins. a-d)** Sub-correlation between ApexMIC's predicted scores and tested MICs on different pathogens.

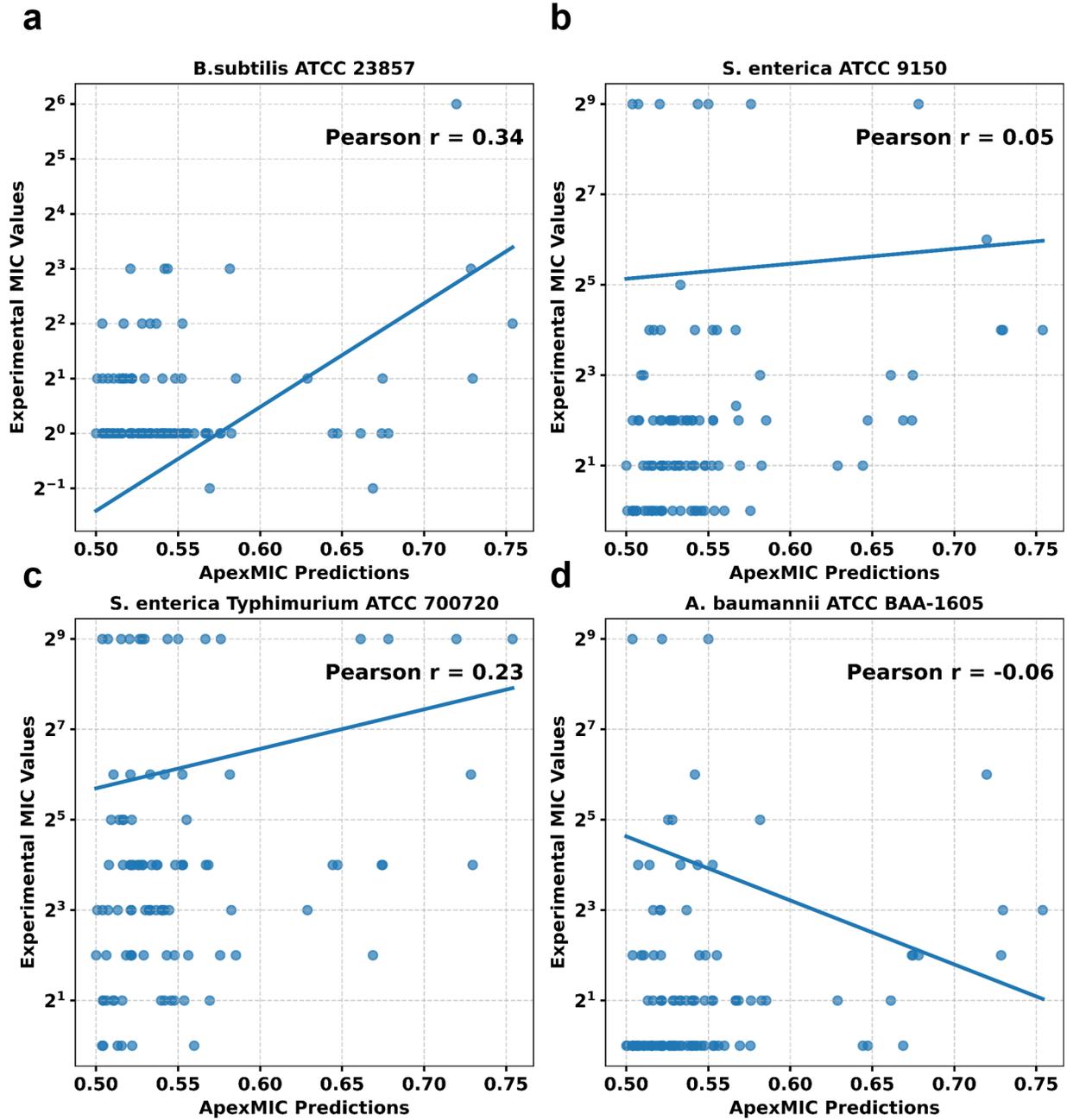

**Supplementary Figure 12. Correlation analysis between ApexMIC's prediction and wet-lab experimental MIC values of amphionins. a-d)** Sub-correlation between ApexMIC's predicted scores and tested MICs on different pathogens.

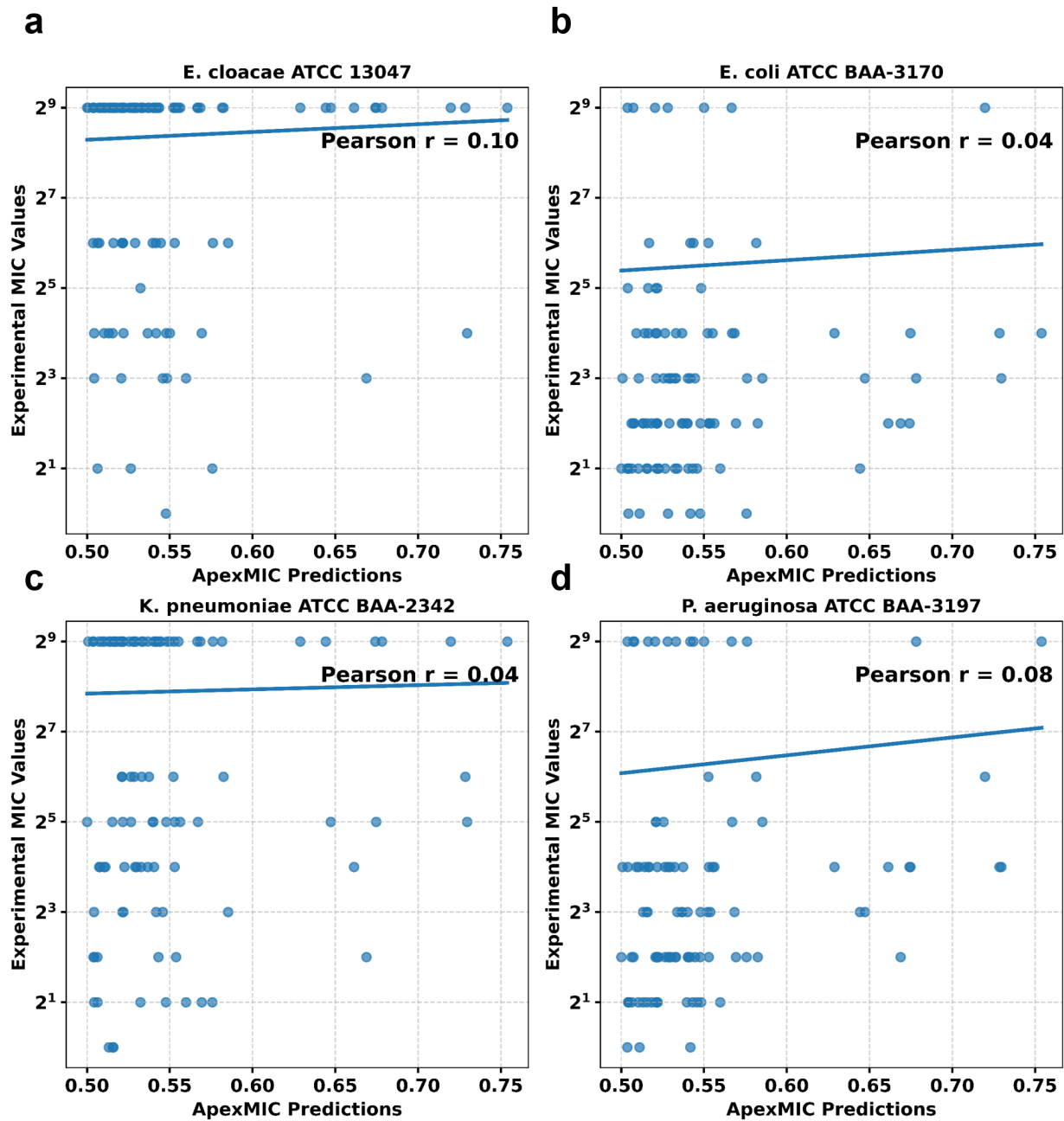

**Supplementary Figure 13. Correlation analysis between ApexMIC's prediction and wet-lab experimental MIC values of amphionins. a-d)** Sub-correlation between ApexMIC's predicted scores and tested MICs on different pathogens.